\documentclass[letterpaper]{article} 
\usepackage[submission]{aaai2026}  
\usepackage{times}  
\usepackage{helvet}  
\usepackage{courier}  
\usepackage[hyphens]{url}  
\usepackage{graphicx} 
\usepackage{natbib}  
\usepackage{caption} 
\usepackage{algorithm}
\usepackage{algorithmic}
\usepackage{newfloat}
\usepackage{listings}
\DeclareCaptionStyle{ruled}{labelfont=normalfont,labelsep=colon,strut=off} 
\floatstyle{ruled}
\newfloat{listing}{tb}{lst}{}
\floatname{listing}{Listing}
\usepackage[table,dvipsnames]{xcolor}
\usepackage{graphicx}
\usepackage{subcaption}
\usepackage{booktabs}
\usepackage{lipsum}
\usepackage{natbib}
\usepackage{verbatimbox}
\usepackage{url}
\usepackage{amsmath,amssymb}
\usepackage[capitalize]{cleveref}
\usepackage{multirow}
\usepackage{pifont}
\usepackage[dvipsnames]{xcolor}
\usepackage{enumitem}

\newcommand{\xmark}{\large{\textcolor{BrickRed}{\text{\sffamily X}}}}%
\newcommand{\cmark}{\large{\textcolor{ForestGreen}{\textbf{\checkmark}}}}

\title{SCRum-9: Multilingual Stance Classification over Rumours on Social Media}
\author{
    Yue Li,
    Jake Vasilakes,
    Zhixue Zhao,
    Carolina Scarton
}
\affiliations{
    \textsuperscript{\rm 1}Department of Computer Science, University of Sheffield, UK\\
        \texttt{\{yli381,j.vasilakes,zhixue.zhao,c.scarton\}@sheffield.ac.uk}\\

}

\usepackage{bibentry}

\begin{document}

\maketitle

\begin{abstract}

We introduce \textbf{SCRum-9}, the largest multilingual \textbf{S}tance \textbf{C}lassification dataset for \textbf{Rum}our analysis in \textbf{9} languages, containing 7,516 tweets from X. SCRum-9 goes beyond existing stance classification datasets by covering more languages, linking examples to more fact-checked claims (2.1k), and including confidence-related annotations from multiple annotators to account for intra- and inter-annotator variability (Figure \ref{fig:teaser}). Annotations were made by at least two native speakers per language, totalling more than 405 hours of annotation and 8,150 dollars in compensation. Further, SCRum-9 is used to benchmark five large language models (LLMs) and two multilingual masked language models (MLMs) in In-Context Learning (ICL) and fine-tuning setups. This paper also innovates by exploring the use of multilingual synthetic data for rumour stance classification, showing that even LLMs with weak ICL performance can produce valuable synthetic data for fine-tuning small MLMs, enabling them to achieve higher performance than zero-shot ICL in LLMs.
Finally, we examine the relationship between model predictions and human uncertainty on ambiguous cases finding that model predictions often match the second-choice labels assigned by annotators, rather than diverging entirely from human judgments. SCRum-9 is publicly released to the research community with potential to foster further research on multilingual analysis of misleading narratives on social media.
\end{abstract}


\section{Introduction}

Social media amplifies the spread of information, including dis- or misinformation and rumours, which often cross linguistic and geographic boundaries~\cite{shu2020combating}. Rumour stance classification \cite{zubiaga-etal-2018-survey} plays a key role in rumour analysis on social media by identifying whether users support, refute, or question a rumour. However, most existing datasets \citep{derczynski-etal-2017-semeval,gorrell-etal-2019-semeval} focus exclusively on English, despite the multilingual nature of online rumours. Efforts have been made to construct multilingual datasets for generic stance classification \citep{zotova-etal-2020-multilingual,vamvas2020xstance,espana-bonet-2023-multilingual,agerri2021vaxxstance,hamdi2021multilingual}, whose task framing, label granularity, and evaluation protocol are fundamentally different from stance classification for rumour or misinformation analysis \citep{hardalov-etal-2022-survey,scarton-etal-2020-measuring}. 

\begin{figure}[ht]
    \centering
    \includegraphics[width=\linewidth]{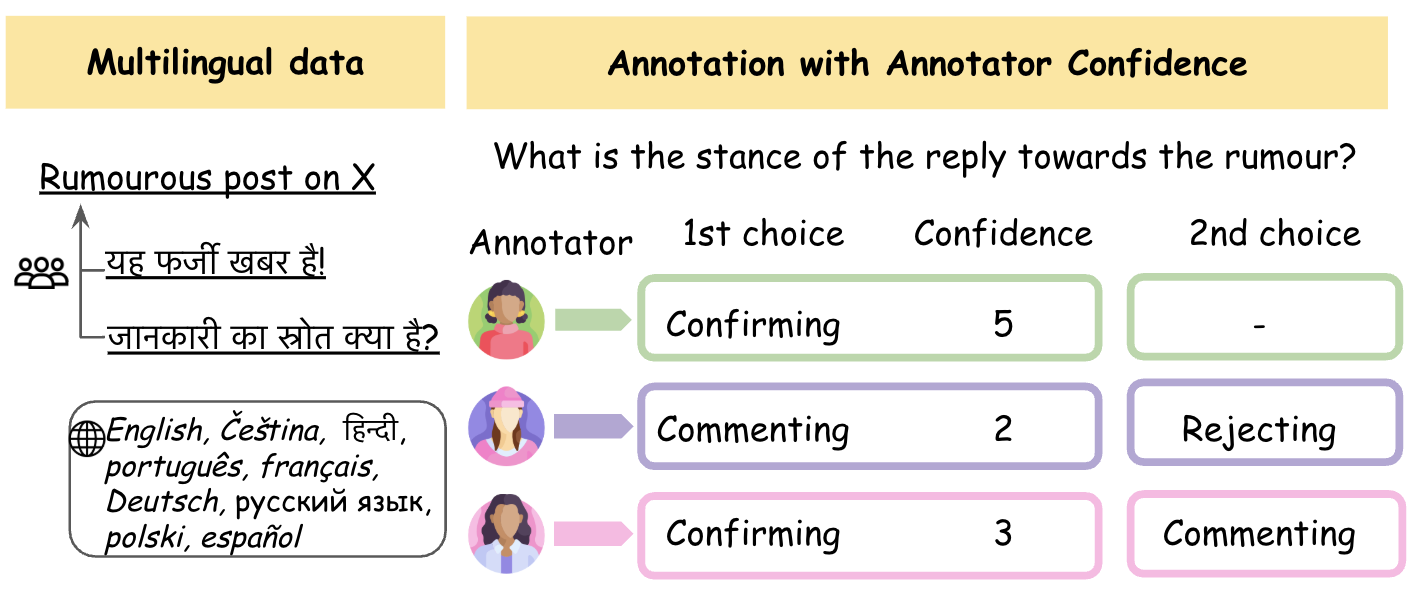}
    \caption{An illustration of the multilinguality and annotation design of SCRum-9.}
    \label{fig:teaser}
\end{figure}

This paper introduces SCRum-9, the first large-scale multilingual benchmark dataset for rumour stance classification. The dataset consists of 7,516 reply tweets across nine diverse languages: Czech (CS), German (DE), English (EN), Spanish (ES), French (FR), Hindi (HI), Polish (PL), Portuguese (PT), and Russian (RU). SCRum-9 covers a broad range of topics with 2,156 distinct rumours collected from fact-checking websites. Each reply is manually annotated using established rumour stance classification schemes \citep{gorrell-etal-2019-semeval}, with an additional design (Figure \ref{fig:teaser}) allowing annotators to provide a second-choice label when unsure, which aims to model annotator uncertainty \citep{mu2023vaxxhesitancy}. To the best of our knowledge, SCRum-9 is the most topically and linguistically diverse rumour stance classification dataset to date\footnote{To be released under a CC BY-NC-SA 4.0 license upon acceptance.}. We also release all raw annotations to facilitate studies on annotators' uncertainty and subjectivity in multilingual stance classification.

\begin{table*}[t!]
\centering
\scalebox{0.85}{
\begin{tabular}{l|cccccc}
\toprule
\multirow{2}{5em}{Dataset} & \multirow{2}{2em}{Size} & \multirow{2}{3em}{\#Stance} & \multirow{2}{3em}{\#Lang} & \#Fact- & Confidence & \multirow{2}{10em}{Released Annotations}\\
 &  &  &  & Checks  & Annotation  &  \\
\midrule
    
PHEME \citep{zubiaga2016analysing}  & 5.8k  & 4  & 2  & 5  & \xmark & Aggregated\\
Stanceosaurus \citep{zheng-etal-2022-stanceosaurus}  & 28k  & 5  & 3 & 250  & \xmark & Aggregated\\
Stanceosaurus 2.0 \citep{lavrouk-etal-2024-stanceosaurus} & 32k  & 5   & 2  & 291  & \xmark  & Aggregated\\
\midrule
\textbf{SCRum-9 (Ours)} & \textbf{7.5k}  & \textbf{4}  & \textbf{9} & \textbf{2,156} & \cmark & \textbf{Raw \& Aggregated}\\
\bottomrule
\end{tabular}
}
\caption{Comparison between SCRum-9 and existing multilingual stance classification datasets for misinformation or rumour analysis.}
\label{tab:Datasets}
\end{table*}

We benchmark five open-source multilingual LLMs using ICL and observe substantial performance disparities across English and relatively low-resource languages included in SCRum-9. We conduct extensive analyses of strategies for improving ICL performance in non-English, including (1) direct machine translating non-English target input into English, or English demonstration examples into target non-English; and (2) incorporating language alignment signals into the prompt. We also evaluate two MLMs fine-tuned on English training data, machine-translated multilingual data, or synthetic multilingual data generated by LLMs. 

Our key findings are as follows:

\begin{itemize}

    \item Machine translating target input or demonstration examples is an effective and competitive strategy to improve ICL performance on relatively high-resource languages whose translation quality is reliable. 

    \item Zero-shot ICL significantly outperforms or matches the performance of MLMs fine-tuned with English or machine-translated multilingual data.

    \item MLMs fine-tuned with LLM-generated multilingual data demonstrate promising results which are comparable to or even better than zero-shot ICL performance with the same LLM, while requiring substantially lower computational costs.

    \item In ambiguous cases where annotators provide second-choice labels, model predictions sometimes align with these alternatives, suggesting that the outputs mirror human uncertainty rather than being purely errors.

\end{itemize}

\section{Related Work}

\subsection{Multilingual Stance Classification Datasets} 

Efforts have been made to construct multilingual stance classification datasets, such as CIC \citep{zotova-etal-2020-multilingual}, Xstance \citep{vamvas2020xstance}, PoliOscar \citep{espana-bonet-2023-multilingual}, VaxxStance \citep{agerri2021vaxxstance} and NEWSEYE \citep{hamdi2021multilingual}. However, these datasets are typically centred around political opinions or social issues, rather than for misinformation or rumour analysis. Stance classification for the latter purpose, such as rumour stance classification, is often framed with distinct and more nuanced stance categories to support fact-checking or rumour verification (such as the \textit{questioning} stance to identify whether a related information is asked for verification) \citep{zubiaga-etal-2018-survey,hardalov2022survey}. The differences in task framing, label granularity, and annotation guidelines render those general stance classification datasets not applicable to model development and evaluation in disinformation or rumour analysis scenarios. 

\paragraph{Language Coverage} There are only three multilingual stance classification datasets established for disinformation or rumour analysis as presented in Table \ref{tab:Datasets}, including PHEME  (English and German), Stanceosaurus (English, Hindi, and Arabic) and Stanceosaurus 2.0 (Russian and Spanish), covering five non-English languages in total. Our proposed SCRum-9 dataset expands the linguistic coverage by incorporating four additional non-English languages not present in the above previous work: Polish, Czech, French, and Portuguese. 

\paragraph{Annotation Protocol} Stance classification is inherently subjective, as annotators may interpret the same post differently due to linguistic ambiguity, different cultural background, or personal judgments \citep{mu2023vaxxhesitancy,wu2023don}. Existing English \citep{derczynski-etal-2017-semeval,gorrell2019semeval} and multilingual \citep{zubiaga2016analysing} rumour stance classification datasets release the majority-voted labels among multiple annotators as the final gold standard labels. However, the RumourEval 2017 dataset contains highly similar tweets with different aggregated stance labels \citep{derczynski-etal-2017-semeval,li-etal-2019-eventai,garcia-lozano-etal-2017-mama}, illustrating inconsistencies and limitation that arise from relying solely on majority voting. Different from prior work, our annotation protocol explicitly encourages annotators to indicate their confidence level and provide a secondary label when uncertain. We release all raw annotation data from each annotator to enable future research on annotation uncertainty and subjectivity.

\subsection{Multilingual Rumour Stance Classification} 

Several studies have explored multilingual or cross-lingual stance classification \citep{hardalov2022few,zhang2023cross,scarton2021cross,zheng-etal-2022-stanceosaurus,barriere2022cofe}, but they mainly focus on general stance classification, which differs significantly in task framing and evaluation protocol from rumour stance classification \citep{scarton2020measuring}. Furthermore, prior work has experimented with multilingual MLMs (e.g., multilingual BERT \citep{scarton2021cross}), with the focus on transferring knowledge learnt from the English source stance classification dataset to the target-language dataset. However, current LLMs~\citep{grattafiori2024llama,le2023bloom} have offered support for many relatively high-resource languages (e.g., the eight non-English languages in SCRum-9). The multilingual capability of LLMs on stance classification, especially rumour stance classification, and their potential to generate non-English synthetic data for those languages in multilingual learning remain under-explored.

\section{The SCRum-9 Multilingual Dataset} 

We construct our dataset from rumours collected on fact-checking websites. To ensure both topic consistency and diversity across languages, we apply a topic-based filtering strategy when selecting rumours for annotation. Our annotation scheme follows PHEME and RumourEval datasets \citep{derczynski-etal-2017-semeval,gorrell-etal-2019-semeval}, while introducing additional design to explicitly capture annotator uncertainty inspired by \citet{mu2023vaxxhesitancy}. Notably, SCRum-9 is different from existing rumour stance classification datasets that either focus on emerging rumours not known a priori (e.g., PHEME and RumourEval 2017 \citep{derczynski-etal-2017-semeval}) or restrict to a single topic (e.g., natural disasters in RumourEval 2019 \citep{gorrell-etal-2019-semeval}).

\subsection{Data Collection}
\label{sec:data_collection}

We collect fact-checked claims and corresponding X posts from two sources:

\begin{itemize}
     
\item \textbf{The Database of Known Fakes (DBKF):} DBKF is a publicly available database of existing fact-checks and associated metadata from trusted fact-checking organisations.\footnote{\url{https://dbkf.ontotext.com}} It contains fact-checked claims, links to the fact-checking articles, as well as links to news and social media posts related to the claim, including those from X. The fact-checks included in DBKF cover a wide range of languages, including those listed above. We query DBKF for all data points in one of the nine target languages that are linked to one or more X posts.

\item \textbf{Fact-Checking Websites:} We identify different trusted fact-checking websites for each language (see \cref{tab:data_collection_sources} in the Appendix for specific websites). With permission, we develop web scraping tools that crawl each site, identify all fact-check articles, and extract the claim and all associated X links.

\end{itemize}

The result of this step is a collection of claims linked to X tweet URLs. We use the X API\footnote{\url{https://developer.x.com/en/docs/x-api}} to obtain each tweet’s text, replies, and other metadata. Given the large number of replies that many posts have, obtaining all replies would quickly reach the 1 million post limit enforced by the X API. We opt to only collect the tweets directly replying to the rumourous source tweet, as the responses in extended conversations become less informative and/or gradually shift towards topics less relevant to the rumour \citep{kochkina-liakata-2020-estimating}. We sample up to 60 direct replies for each post, keeping posts that attract sufficient public engagement (i.e., with more than 25 replies) and informative replies that do not contain only URLs, user mentions, or emojis.

\subsection{Topic-Based Tweet Filtering and Pre-Processing}
\label{sec:filtering}

To ensure topic diversity but also mitigate models leveraging event-specific biases in model evaluation, we employed the multilingual BERTopic model \citep{grootendorst2022bertopicneuraltopicmodeling}, alongside manual review, to identify a set of topics that occur across languages, resulting in 63 coherent topics  (details can be found in the Appendix). The topics range from broad ones such as COVID-19 and natural disasters, to specific rumours, e.g., that Switzerland has outlawed mammograms. We keep only rumourous source tweets that are assigned to these 63 topics by the topic model, except for Hindi, Czech, and Russian, where filtering using the topic model would severely reduce the total number of examples. After filtering, 53 of the original 63 topics were represented.  Finally, we sample up to 1,500 tweet-reply pairs per language for annotation such that we maximise the total number of topics represented.
Statistics of the source tweets, replies, and fact-checked claims that were collected, filtered, and annotated are presented in Table \ref{tab:collected_tweet_stats} in the Appendix.

Before annotation, the tweet-reply pairs were anonymised by replacing all user \texttt{@} mentions with the string \texttt{@USER} and all URLs within tweets were replaced with \texttt{HTTPURL}. We discuss ethical considerations related to data anonymity in the Appendix.

\subsection{Data Annotation}

The annotation task is to determine the stance of a reply tweet towards its rumourous source tweet within a conversation thread on X. Stance is defined as the way in which the reply tweet's author regards the source tweet. Following PHEME and RumourEval datasets \citep{gorrell-etal-2019-semeval}, there are four possible stances: 

\begin{itemize}
    \item \textbf{Confirming} - The author expresses agreement with source tweet.
    \item \textbf{Rejecting} - The author expresses disagreement with or denial of the source tweet. 
    \item \textbf{Questioning} - The author asks for additional evidence or confirmation of the source tweet. 
    \item \textbf{Commenting} - The author comments on the source tweet but does not take a clear stance. This includes cases in which the reply is unrelated to the source tweet. 
\end{itemize}

Since SCRum-9 focuses on text-based rumour stance classification, annotators are also instructed to label replies that rely exclusively on the attached images or videos to interpret the stance as \textit{Only refers to image/video}. These instances are excluded from the final dataset.

\paragraph{Confidence-Based Annotation} Annotators are instructed to provide one or two stance annotations for each tweet-reply pair, based on their annotation confidence. Specifically, each annotator indicates a first-choice stance label, accompanied by a confidence rating on a 5-point Likert scale (1 = extremely uncertain, 5 = absolutely certain). If the confidence rating is lower than three, annotators are required to provide a second-choice stance label. To account for cases where no specific second label can be reasonably assigned, we include \textit{Highly Uncertain} as an additional option. Detailed annotation guidelines are provided in the Appendix.

\subsubsection{Annotator Recruitment and Quality Control}

Annotators are native speakers of each target language from universities and research institutions in our network. The native English speakers were predominantly from the UK, and the native Portuguese speakers were from Brazil. To comply with our ethical policy, annotators received an information sheet, outlining the details regarding their participation and needed to sign a consent form. 

Participants received a 40-minute online training session prior to beginning the annotation task, where we introduced the annotation guidelines, process, and interface. After the training session, participants completed quality control annotations on 60 held-out examples in English, which had been annotated by three researchers experts in the task. We selected participants who reached a high level of agreement with the researchers' annotations.\footnote{We did not set a hard agreement threshold, instead evaluating each individual annotator in the context of the others, per language.} Annotators were compensated at the rate of \$20 (USD) per hour (or equivalent in their currency of choice) in the form of an online gift voucher (e.g. Amazon voucher). The total annotation cost was approximately 8,150 dollars.

\paragraph{Annotation Process} 

The main annotation process was split into two rounds. In the first round, two annotators were assigned to each tweet-reply pair. In the second round, we collected all tweet-reply pairs on which the two annotators disagreed in their first-choice label during the first round and assigned each of these examples to a third annotator in an effort to obtain a consensus. Thus, each example in the dataset contains a first-choice stance annotation, a confidence score, and a possible second-choice stance annotation from two or three annotators. Annotators were assigned to examples using the EffiARA library
\cite{cook-etal-2025-efficient}. The open-source GATE Teamware tool\footnote{\url{https://annotate.gate.ac.uk/}} \cite{wilby-etal-2023-gate} is used for conducting the annotation (see \cref{fig:gate_screenshot} in the Appendix for a screenshot of the annotation interface).

\subsection{Dataset Overview}

\subsubsection{Label Aggregation}

Since each annotator provides multiple signals (i.e., first-choice label, confidence and possible second-choice label), the annotations can be transformed and aggregated in different ways for use in model evaluation. Therefore, we explore and implement eight aggregation methods, including four hard-label-based (e.g., majority voting with or without considering second-choice labels) and four soft-label-based (e.g, Bayesian soft voting \citep{wu-etal-2023-dont}) methods. We then compute the cosine agreement between the aggregated labels obtained using each pair of methods. Full details and agreement scores are provided in the Appendix. The results suggest that overall the agreement between aggregation methods is high, with the lowest agreement being 0.863. Accordingly, we adopt the conventional \textit{majority-voted first-choice label} as the primary aggregation label for model evaluation in this paper. We also conduct evaluations incorporating the second-choice label to explore the relationship between model predictions and annotator uncertainty.

\begin{figure}[h!]
    \centering
    \includegraphics[width=\linewidth]{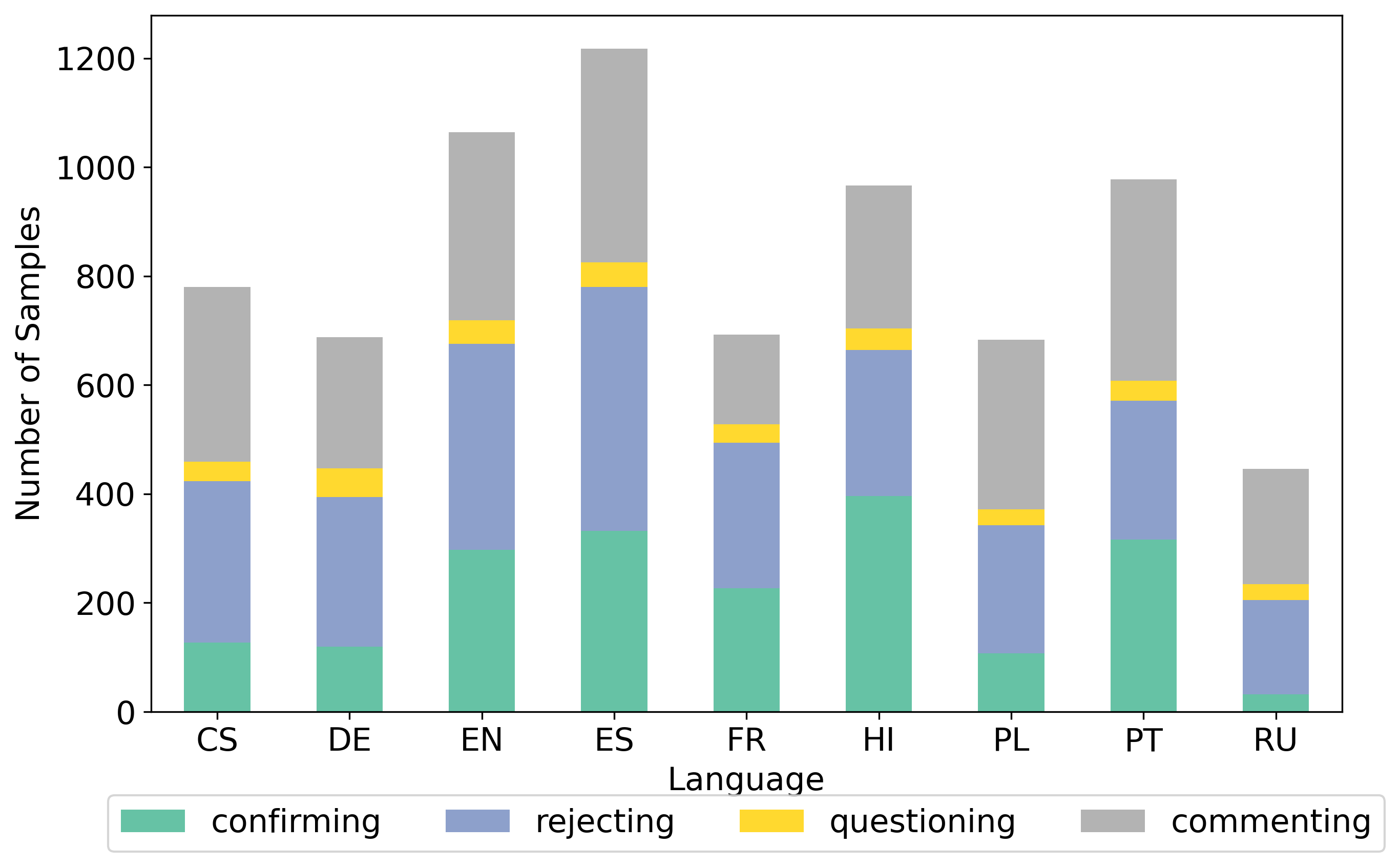}
    \caption{Statistics for SCRum-9 with labels determined with majority-voting over the first-choice stance labels.}
    \label{fig:data_stats}
\end{figure}

\paragraph{Dataset Statistics} The summary statistics of SCRum-9 are presented in Figure \ref{fig:data_stats}, where the labels are determined with majority-voting over the first-choice stances as we discussed above. After removing examples which the majority of annotators labelled as \textit{Only refers to image/video}, SCRum-9 contains 7,516 annotated tweet-reply pairs in total. The number of tweets per language ranges from 446 (in Russian) to 1,218 (in Spanish), with 835 tweets per language on average. The data size is comparable to existing rumour stance classification test sets, such as 1,049 and 1,872 in RumourEval 2017 and 2019 official test sets respectively.
Similar with previous rumour stance classification datasets, the stance class distribution in SCRum-9 is imbalanced, reflecting the natural distribution of stance observed in online rumour-related discussions. The \textit{confirming}, \textit{rejecting}, and \textit{commenting} classes are each relatively evenly represented, accounting for roughly 30\% of the dataset, although this proportion varies across languages. The \textit{questioning} class is less represented in SCRum-9, while it normally exhibits the lowest linguistic diversity among the four classes \citep{li-scarton-2024-identify}.

\subsubsection{Inter-Annotator Agreement}

We calculate the following two kinds of inter-annotator agreement scores:

\begin{itemize}
    \item \textbf{Cosine-based agreement} \citep{dumitrache2018crowdtruth}: We represent the first-choice label, confidence level and second-choice label of each annotator with vectors, and calculate the cosine agreement scores. The vector encodes the first- and second-choice labels weighted with the confidence scores, followed by normalisation. Details of the vector calculation refer to the \textit{MVC2 (Majority Vote with Confidence and Second-choice)} aggregation method in the Appendix.  

    \item \textbf{Percentage-based agreement}: We also report the percentage agreement over the first-choice labels to enable direct comparison with current rumour stance classification datasets (i.e., RumourEval datasets). 
\end{itemize}

\begin{table}[h]
\centering
\scalebox{0.67}{
\begin{tabular}{l|ccccccccc|c}
\toprule

\textbf{Lang} &\textbf{CS}  &\textbf{DE}   &\textbf{EN}   &\textbf{ES}&\textbf{FR}  &\textbf{HI}&\textbf{PL} &\textbf{PT}&\textbf{RU}&\textbf{All} \\
\toprule
\texttt{cosine-1}&0.55&0.38&0.48&0.51&0.52&0.4&0.47&0.52&0.48&0.49\\
\texttt{cosine-2}&0.65&0.55&0.62&0.60&0.65&0.47&0.61&0.59&0.59&0.60\\
\midrule
\texttt{percent}&0.61&0.40&0.59&0.57&0.62&0.52&0.54&0.58&0.55&0.55\\

\bottomrule
\end{tabular}
}
\caption{Average cosine (\texttt{cosine}) and percentage (\texttt{percent}) agreement scores across annotators for each language. We report cosine agreement computed using the first-choice label only (\texttt{cosine-1}) as well as using the first-choice, second-choice and confidence score (\texttt{cosine-2}).}
\label{tab:agreement}
\end{table}

As presented in Table \ref{tab:agreement}, SCRum-9 achieves an average percentage agreement score of 0.55 (ranging from 0.40 on German to 0.62 on French), comparable to current rumour stance classification datasets, where the agreement score on reply tweets reaches 0.62 in RumourEval 2017. Furthermore, when incorporating first-choice and second-choice labels along with confidence scores, the cosine-based agreement increases substantially across languages, highlighting the subjective and uncertain nature of this task. 

We find that out of 18,280 total annotations, 11,744 (64\%) are given a confidence score of 4 or lower, and 4,707 (26\%) are assigned a secondary label. 
Annotator disagreement is most pronounced between the \textit{confirming} and \textit{commenting} categories. We also observe similar pattern on annotators' second-choice labels (as shown in Figure  \ref{fig:first_second_label_cm} in the Appendix). In most of ambiguous cases, annotators select \textit{confirming} or \textit{rejecting} as the first-choice label while considering \textit{commenting} as a possible alternative. The next most frequent pattern is the reverse: annotators select \textit{commenting} as the first-choice label but also consider \textit{confirming} or \textit{rejecting} as plausible options.

\section{Experiments}

In real-world applications, two primary approaches are commonly adopted to perform multilingual classification tasks: applying ICL with multilingual LLMs, or fine-tuning multilingual MLMs on English or multilingual data \citep{li2025s,RazumovskaiaAnalyzing}. In this section, we compare these two approaches across the nine languages in the SCRum-9 dataset. Our experimental setup and analysis focus on: (1) disparities in zero-shot baseline ICL performance across languages and potential strategies to further improve ICL performance, especially on non-English; and (2) the effectiveness of fine-tuning small MLMs with English, translated multilingual data and LLM-generated synthetic multilingual data.

\subsection{Evaluation Settings}

\subsubsection{In-Context Learning with LLMs}

We evaluate five open source instruction-tuned multilingual LLMs: Qwen2.5\footnote{\url{https://huggingface.co/Qwen/Qwen2.5-7B-Instruct}}, Mistral-v0.3\footnote{\url{https://huggingface.co/mistralai/Mistral-7B-Instruct-v0.3}}, Gemma2\footnote{\url{https://huggingface.co/google/gemma-2-9b-it}}, DeepSeek\footnote{\url{https://huggingface.co/deepseek-ai/deepseek-llm-7b-chat}} and Llama3.2\footnote{\url{https://huggingface.co/meta-llama/Llama-3.2-3B-Instruct}}. Due to computational constraints, their medium-sized variants are considered.

\begin{itemize}

    \item \textbf{Zero-Shot ICL}: We directly prompt the LLMs by describing the task with natural language instructions in English following prior work \citep{zhang-etal-2023-dont}. We explore the following three zero-shot ICL settings: 

    \begin{itemize}
        \item \texttt{Baseline}: Input text is in the original target languages.
        
        \item \texttt{translate-input}: If target-language is non-English, we machine-translate the input text into English. In this study, Google translate\footnote{\url{https://translate.google.com/}} is used for machine translation.

        \item \texttt{align example}: Inspired by recent findings on ICL with extremely low-resource languages \citep{li2025s}, for languages other than English, we provide the input text in target languages, along with four unlabelled examples in English and their target-language translations. The examples are randomly sampled from the RumourEval 2019 training set, which is the current largest English rumour stance classification dataset.
    \end{itemize}

    \item \textbf{Few-Shot ICL}: We prompt the LLMs with 4-shot demonstration examples. The demonstrations are randomly sampled from each stance category in the RumourEval 2019 training data. We provide the demonstrations in the following three settings: 

    \begin{itemize}
        \item \texttt{demo-en}: Demonstration examples are in English.

        \item \texttt{demo-translate}: If the target input is not in English, we machine-translate the demonstration examples into the target languages.

        \item \texttt{demo-align}: We provide demonstration examples in both English and their machine translations in the target languages when the input is not in English. Note that this setting is different from \texttt{align-example} in zero-shot ICL, whose alignment examples are not provided with the stance labels.
    \end{itemize}

\end{itemize}

More specifics, including the prompt template and decoding strategy, are provided in the Appendix.

\paragraph{Fine-Tune Multilingual MLMs} We evaluate two different MLMs: XLM-R \citep{conneau-etal-2020-unsupervised} and XLM-T \citep{barbieri-etal-2022-xlm}. The models share the same architecture, pretrained with multilingual corpora covering all the languages in SCRum-9. XLM-T is intentionally pre-trained with multilingual Twitter data, while XLM-R is pretrained with filtered CommonCrawl data \citep{barbieri-etal-2022-xlm}. We experiment with the following three settings. The hyper-parameter tuning and training details are included in the Appendix: 

\begin{itemize}
    \item \texttt{train-en}: Fine-tuning with English data, i.e., the RumourEval 2019 training set.
    
    \item \texttt{train-translate}: Fine-tuning with machine-translated multilingual data. We machine-translate the English RumourEval 2019 into the eight other SCRUM-9 languages.

    \item \texttt{train-synthetic}: Training with multilingual data synthetically generated by multilingual LLMs. We generate the synthetic data with the five LLMs, respectively. Specifically, to ensure each language contains the same number of replies and rumours, we randomly sample 10 rumours for each language from SCRum-9. For each source tweet, we prompt the LLM to generate $N$ different reply tweets for each stance. The stance defined in each prompt is then used as the labels during fine-tuning. Under the consideration of data efficiency, we opt to generate ten replies per stance per source tweet, resulting in 400 tweets per language and 3,600 multilingual tweets in total\footnote{The number of tweets is slightly lower than 3,600 for certain LLMs since they sometimes refuse to generate replies to misinformation.}. Note that the RumourEval training data we use in the above two approaches contains 6,702 tweets. More details of the prompt template is provided in the Appendix.

\end{itemize}

\paragraph{Evaluation Metrics} Our primary evaluation metric is weighted $F_2$ score ($wF2$)\footnote{$wF2$ gives different weights for each stance: \textit{deny} $=$ \textit{support} $= 0.40$, \textit{query} $= 0.15$ and \textit{comment} $= 0.05$.} \citep{scarton-etal-2020-measuring}, which rewards models with high performance on the \textit{confirming} and \textit{rejecting} classes, being more adequate to rumour stance classification. 

\section{Results and Discussions}

\subsection{Baseline Zero-Shot ICL Performance and Inconsistency Across Languages} 

The LLMs exhibit varying performance and cross-lingual consistency when evaluated with baseline zero-shot ICL on SCRum-9. \textbf{Most of the LLMs achieve their best performance on English, except for Gemma}. We present the average and standard deviation of the $wF2$ scores across the nine languages for each LLM in Figure \ref{fig:mean_std_zeroICL} (detailed results are in the Appendix). LLMs shown in the \textit{bottom-right} area (i.e., high mean with low standard deviation) achieve strong and relatively consistent performance across languages, which is a more desirable outcome. Performances of models on \textit{bottom-left} (i.e., low mean and low standard deviation) are consistently weak, while models on \textit{top-right} (i.e., high mean, high standard deviation) are strong overall but uneven across languages.

The results in Figure \ref{fig:mean_std_zeroICL} show that Gemma demonstrates relatively strong and consistent performance across all languages, with the highest $wF2$ of 0.59 on Spanish and the lowest $wF2$ of 0.48 on Hindi, indicating relatively robust cross-lingual generalisation on rumour stance classification. In contrast, Mistral achieves second-best performance on English ($wF2$ as 0.5) among the five LLMs, but its performance on German and Hindi is only 0.33, suggesting substantial variability across languages. Deepseek achieves overall moderate performances with high disparity, while Llama and Qwen show low performances across all languages, suggesting their limited zero-shot ICL capability for this classification task.

\begin{figure}[h!]
    \centering
    \includegraphics[width=0.708\linewidth]{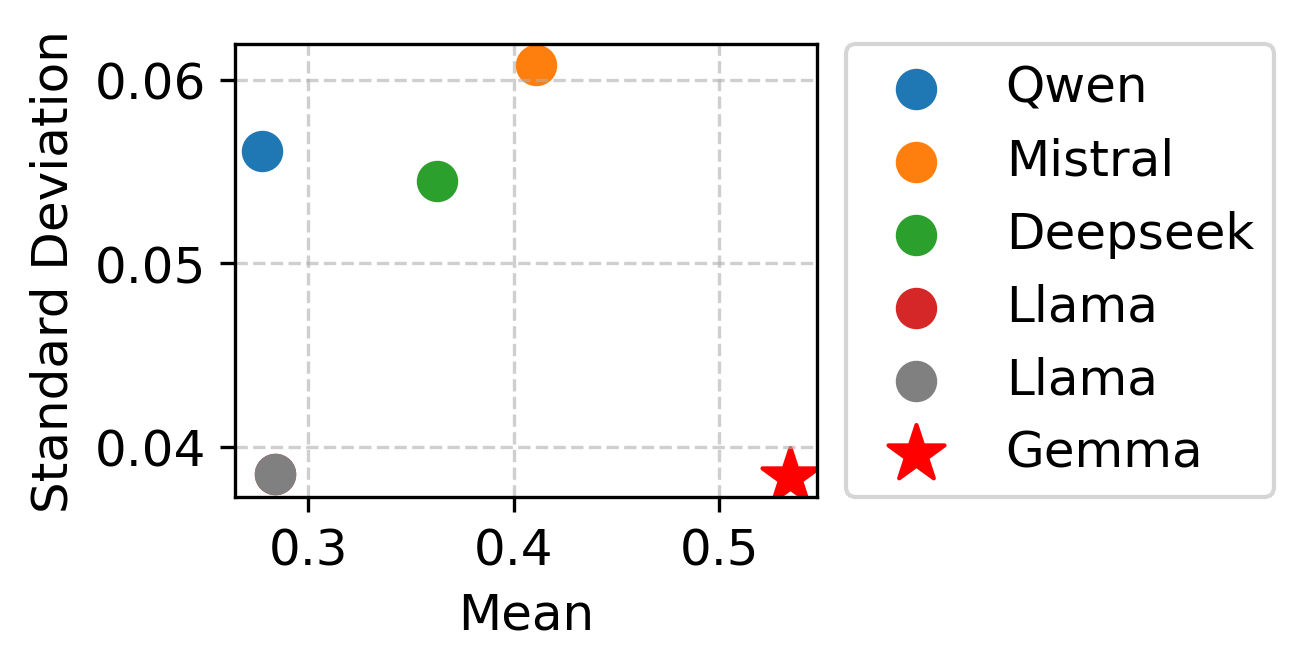}
    \caption{Mean and standard deviation of zero-shot baseline ICL performance ($wF2$) on SCRum-9 with different LLMs. LLMs on \textit{bottom-right} with high mean and low standard deviation exhibit good and relatively consistent performance across languages.}

    \label{fig:mean_std_zeroICL}
\end{figure}

\subsection{ICL for Non-English Rumour Stance Classification: Translation vs. Language Alignment} 

\begin{figure}[h!]
    \centering
    \includegraphics[width=0.9\linewidth]{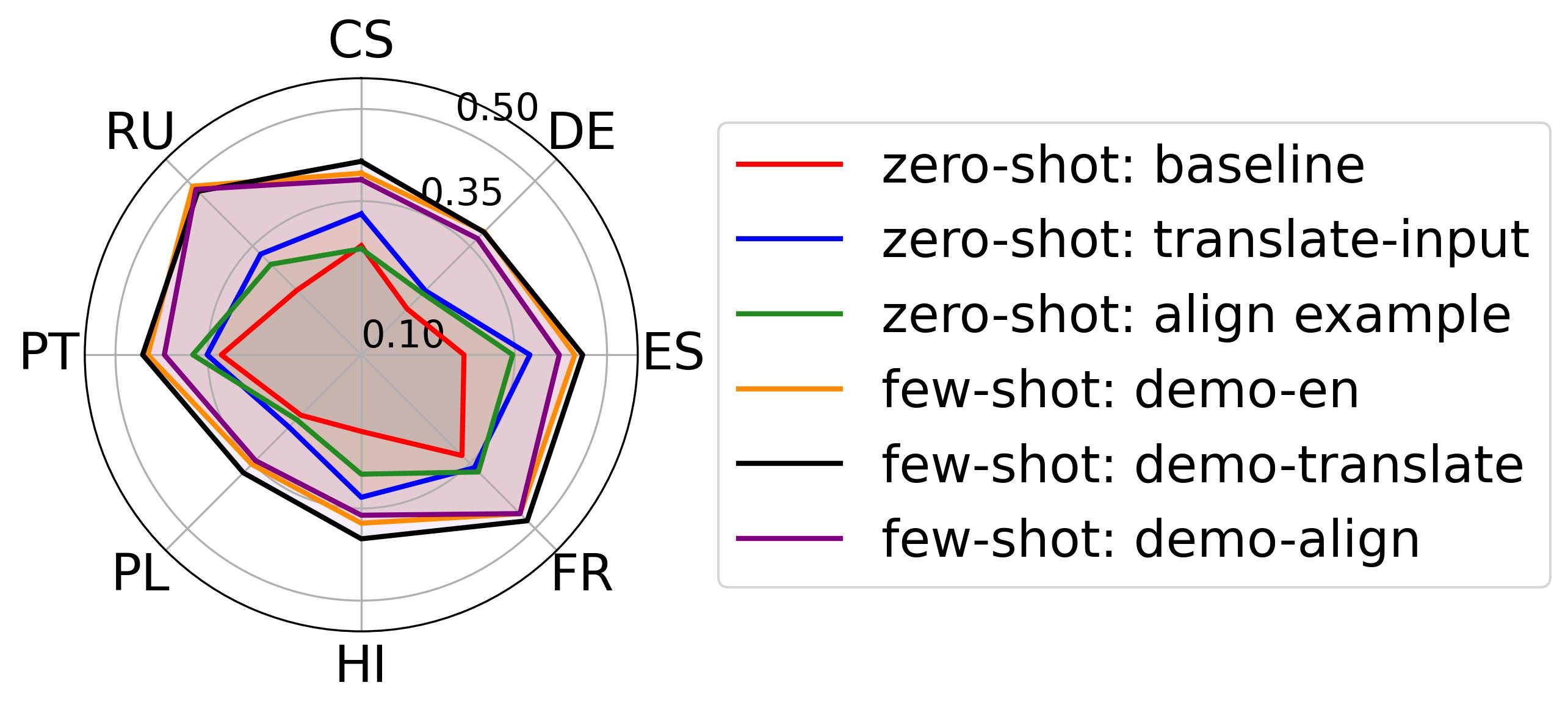}
    \caption{Comparison between ICL performances ($wF2$) across the eight non-English languages with Qwen. }
    \label{fig:non-en-ICL}
\end{figure}

As discussed above, LLMs show performance disparity across languages, typically performing better in English. Therefore, we analyse which approaches are effective to improve the ICL performance for the eight non-English languages in SCRum-9. We present the results on Qwen, as an example, in Figure \ref{fig:non-en-ICL}. Full results of other LLMs can be found in the Appendix.

\paragraph{Zero-Shot ICL} \textbf{Both translating the target input into English and providing unlabelled translation pairs as language alignment signals} generally enhance performance over the baseline zero-shot ICL across all the LLMs and languages. However, we observe that the language alignment strategy might slightly degrade performance in certain cases (e.g., on German with Mistral), consistent with previous findings on extremely low-resources languages \citep{li2025s}. Language alignment also shows no benefit for Llama, which we attribute not only to cross-lingual challenges but also to Llama’s overall weak performance on this task, as indicated in Figure \ref{fig:mean_std_zeroICL}. 

Furthermore, \textbf{for most languages, directly translating the input into English yields stronger improvements}, except for Gemma for which including the language-alignment signal in the prompt consistently produces better results than translation across all the languages. 

\paragraph{Few-Shot ICL} Regardless of whether the demonstration examples are in English or translated into the target non-English, and whether additional language alignment is provided, few-shot ICL generally outperforms zero-shot ICL. This improvement is particularly notable for LLMs such as Llama and Qwen that perform poorly in the zero-shot setting. For example, as shown in Figure \ref{fig:non-en-ICL}, Qwen shows a clear gap between zero-shot and few-shot ICL performances across all the languages. Also, \textbf{in general, providing demonstration examples directly in the target non-English language yields greater gains}. However, Gemma benefits more when using English-only demonstrations for half of the eight non-English languages.

\paragraph{Key Findings} We summarise our two key findings for multilingual rumour stance classification with ICL:

\begin{itemize}

    \item Since the non-English languages in SCRum-9 are relatively high-resource and well supported by machine translation tools (e.g., Google Translate, which we use in this study), \textbf{machine translation generally serves as an effective method for improving ICL performance in these languages}.

    \item Few-shot ICL, even with English demonstration examples, outperforms zero-shot ICL (regardless of whether translation or language alignment strategies are applied), suggesting that \textbf{multilingual LLMs are capable of effectively transferring knowledge across languages in ICL, especially the relatively high-resource languages for rumour stance classification}. 
    
\end{itemize}

\subsection{Cross-Lingual and Multilingual Fine-Tuning MLMs vs. Prompting LLMs} 

Although ICL demonstrates promising performance across languages as we discussed above, it is usually computationally more expensive than fine-tuning and then inferencing with smaller multilingual MLMs. We compare the baseline zero-shot ICL and best ICL performance after improvement of Gemma and Llama (i.e., the best and worst LLMs in baseline ICL as shown in Figure \ref{fig:mean_std_zeroICL}) against XLM-R fine-tuned with either English RumourEval data or its machine-translated multilingual version. The results on XLM-T are similar, which can be found in the Appendix.

As shown in Figure \ref{fig:ICL_FT}, we observe \textbf{a notable gap between Gemma's ICL performance and that of fine-tuned XLM-R across all the languages}. The performance of fine-tuned XLM-R is comparable to Llama's baseline zero-shot ICL; however, providing only four demonstration examples in the prompt for Llama (e.g., Llama(best) in \ref{fig:ICL_FT}) elevates Llama’s performance well beyond that of the fine-tuned XLM-R, resulting in a substantial performance gap.

\begin{figure}[h!]
    \centering
    \includegraphics[width=0.9\linewidth]{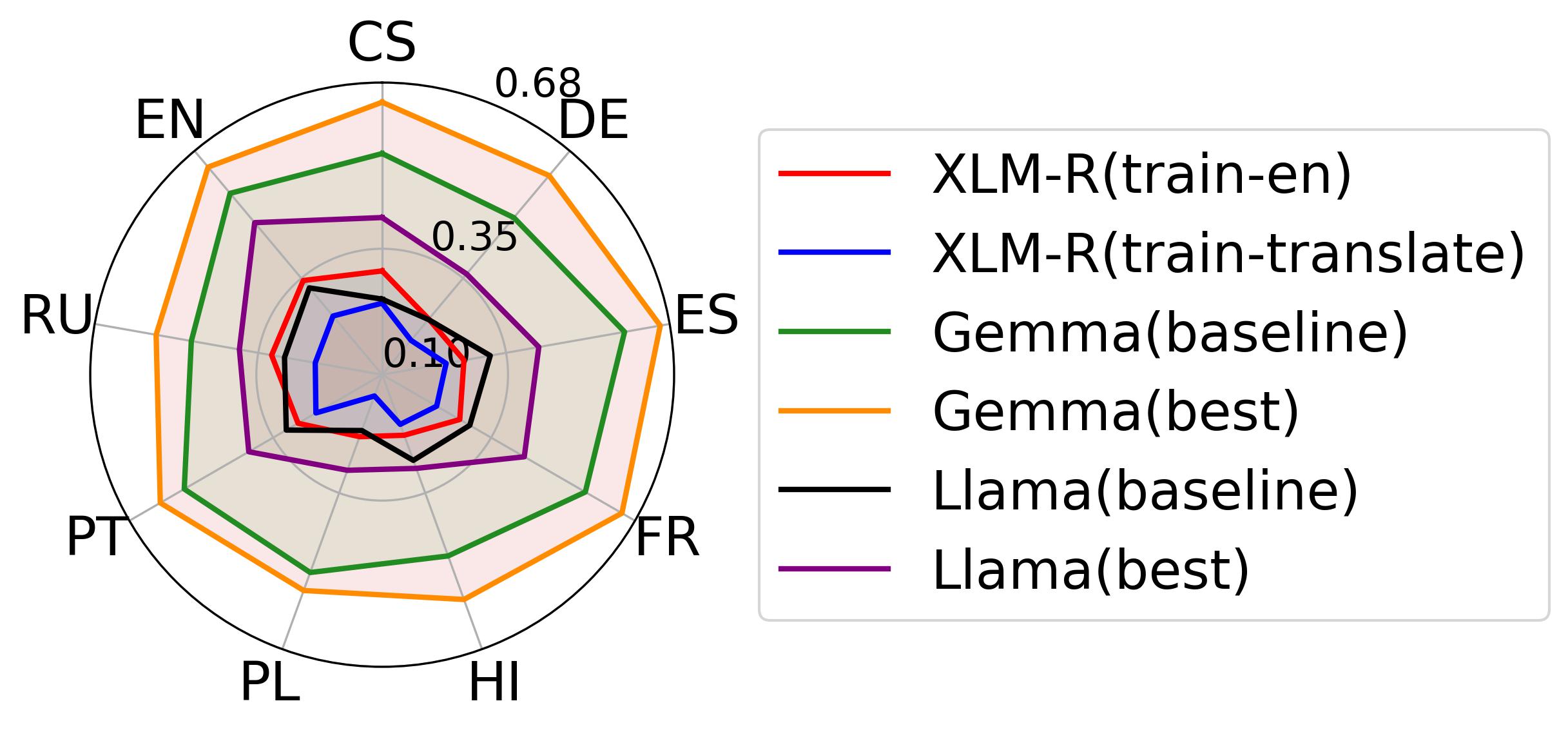}
    \caption{Performance ($wF2$) comparison across languages between (1) XLM-R fine-tuned with English; (2) XLM-R fine-tuned with translated multilingual data; (3) Baseline zero-shot ICL performances of Gemma and Llama; and (4) Best ICL performances of Gemma and Llama.}
    \label{fig:ICL_FT}
\end{figure}

\subsection{Effectiveness of Multilingual Synthetic Data} 

In practice, obtaining multilingual training data is both costly and often impractical, as data collection and human annotation require substantial resources. Consequently, a common approach is cross-lingual transfer, where multilingual MLM is fine-tuned on English data. However, as shown previously, this produces suboptimal performance, substantially lagging behind ICL. Meanwhile, deploying LLMs for inference in rumour stance classification is not always feasible due to computational and financial constraints. A promising alternative, when direct LLM inference is not feasible, is to leverage LLMs to generate multilingual synthetic data, which can subsequently be used to fine-tune MLMs for deployment.

\begin{figure}[h!]
    \centering
    \includegraphics[width=0.8\linewidth]{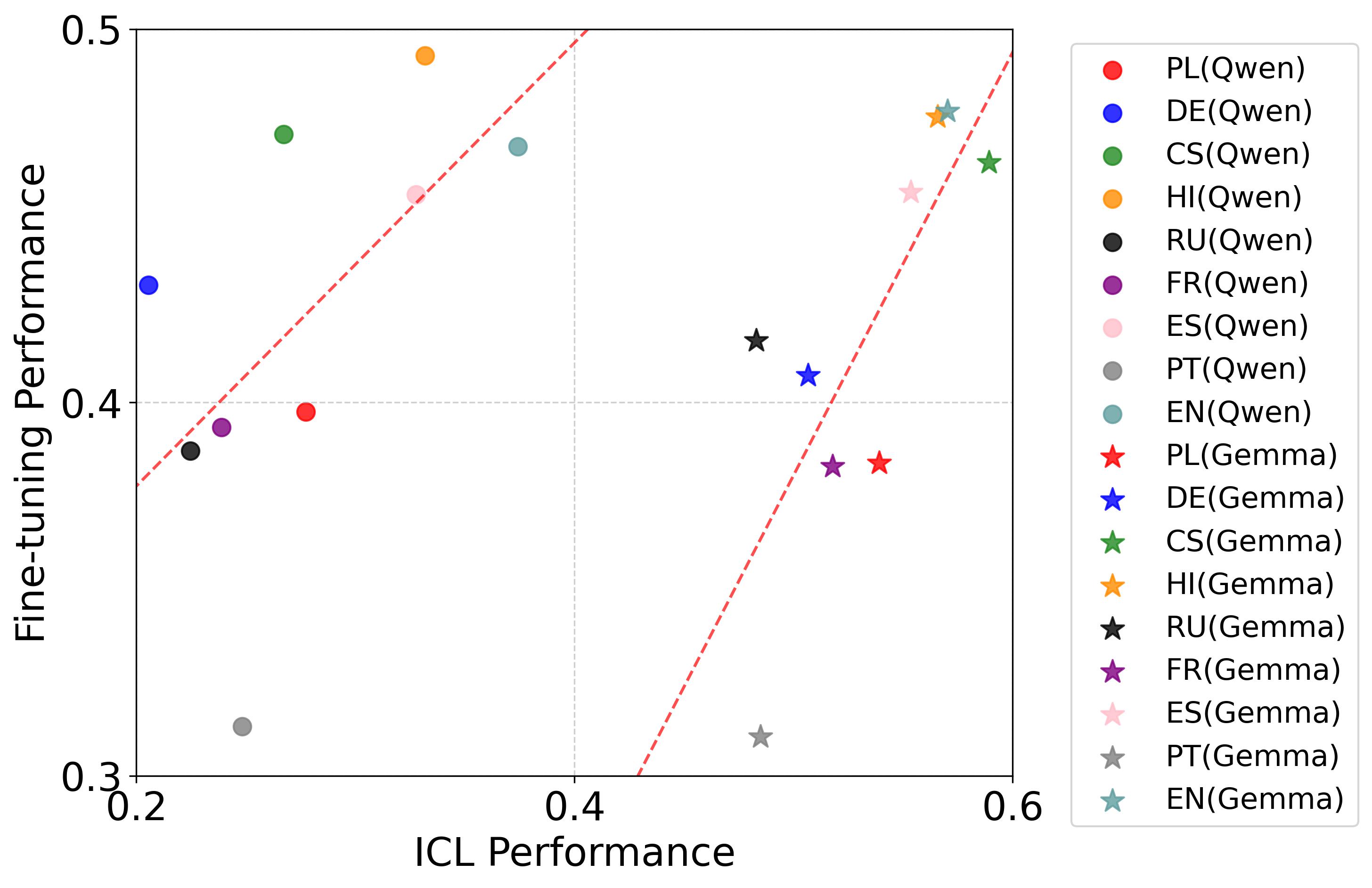}
    \caption{Comparison of Gemma/Qwen baseline zero-shot ICL performance with XLM-R fine-tuned on their generated synthetic data. Gemma is represented by the stars, and Qwen is represented by the circles.}
    \label{fig:ICL_syn}
\end{figure}

\paragraph{Overall Performance} We find that \textbf{fine-tuning MLMs on LLM-generated multilingual synthetic data consistently outperforms training on real-world English data or machine-translated multilingual data}, despite using nearly almost 3,000 fewer training examples. Also, such fine-tuning achieves performance comparable to, or even exceeding, the zero-shot performance of the same LLMs that produce the synthetic data. In particular, for Qwen, Llama, and DeepSeek -- the three LLMs with the weakest baseline zero-shot performance on average (Figure \ref{fig:mean_std_zeroICL}) -- fine-tuning MLMs on their synthetic multilingual data yields significant better performance than ICL. However, for Mistral and Gemma that exhibit the strongest zero-shot baselines on average, fine-tuning with synthetic data might still underperform zero-shot ICL, especially for Gemma. Nonetheless, the performance remains comparable. Full results can be found in the Appendix.

\begin{figure*}[ht!]
\centering
\begin{subfigure}{0.2\textwidth}
\includegraphics[width=1\linewidth]{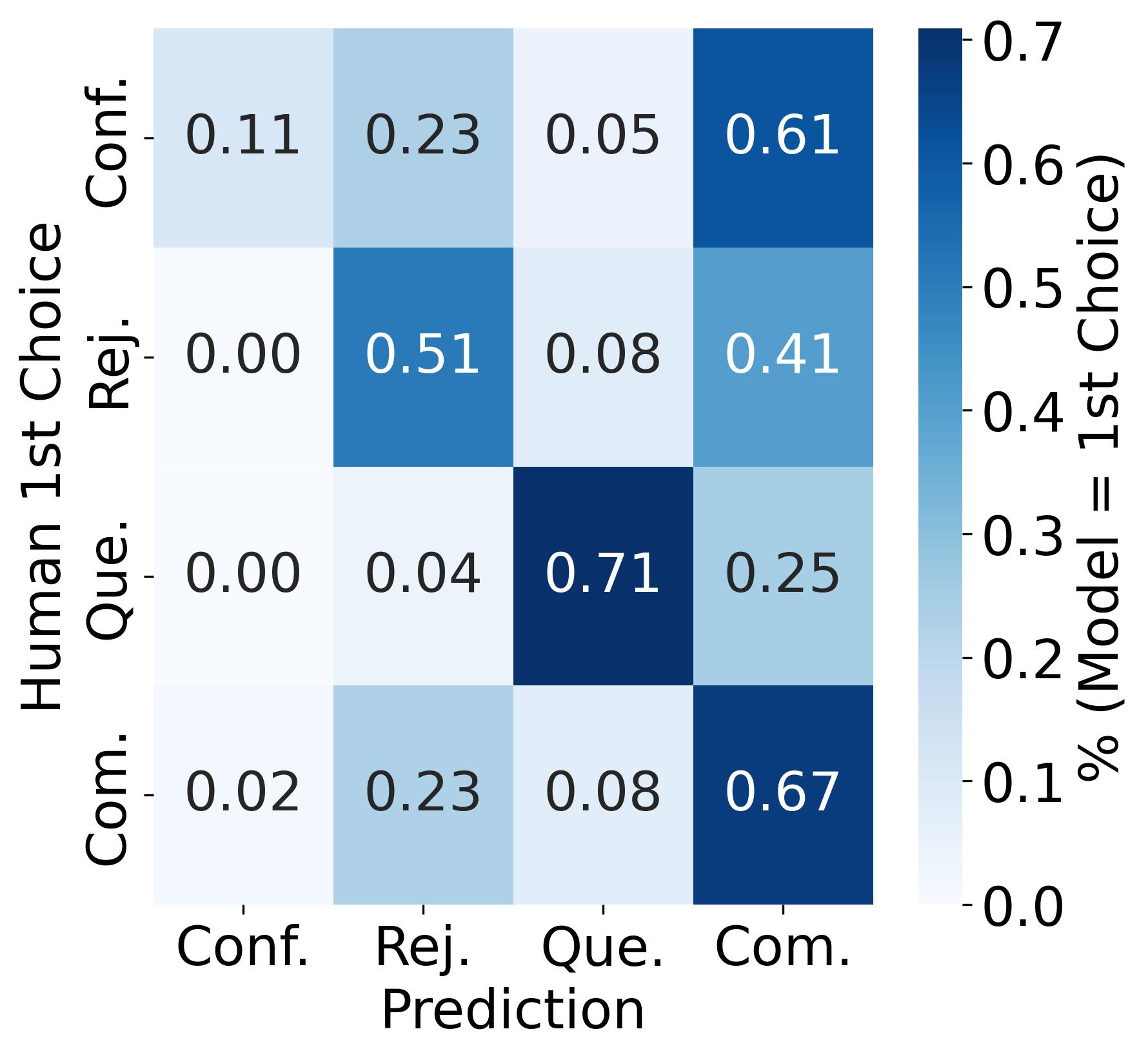}
\caption{Baseline Zero-Shot ICL (1st Choice)}\label{fig:baseline-1}
\end{subfigure}
\hfill
\begin{subfigure}{0.2\textwidth}
\includegraphics[width=1\linewidth]{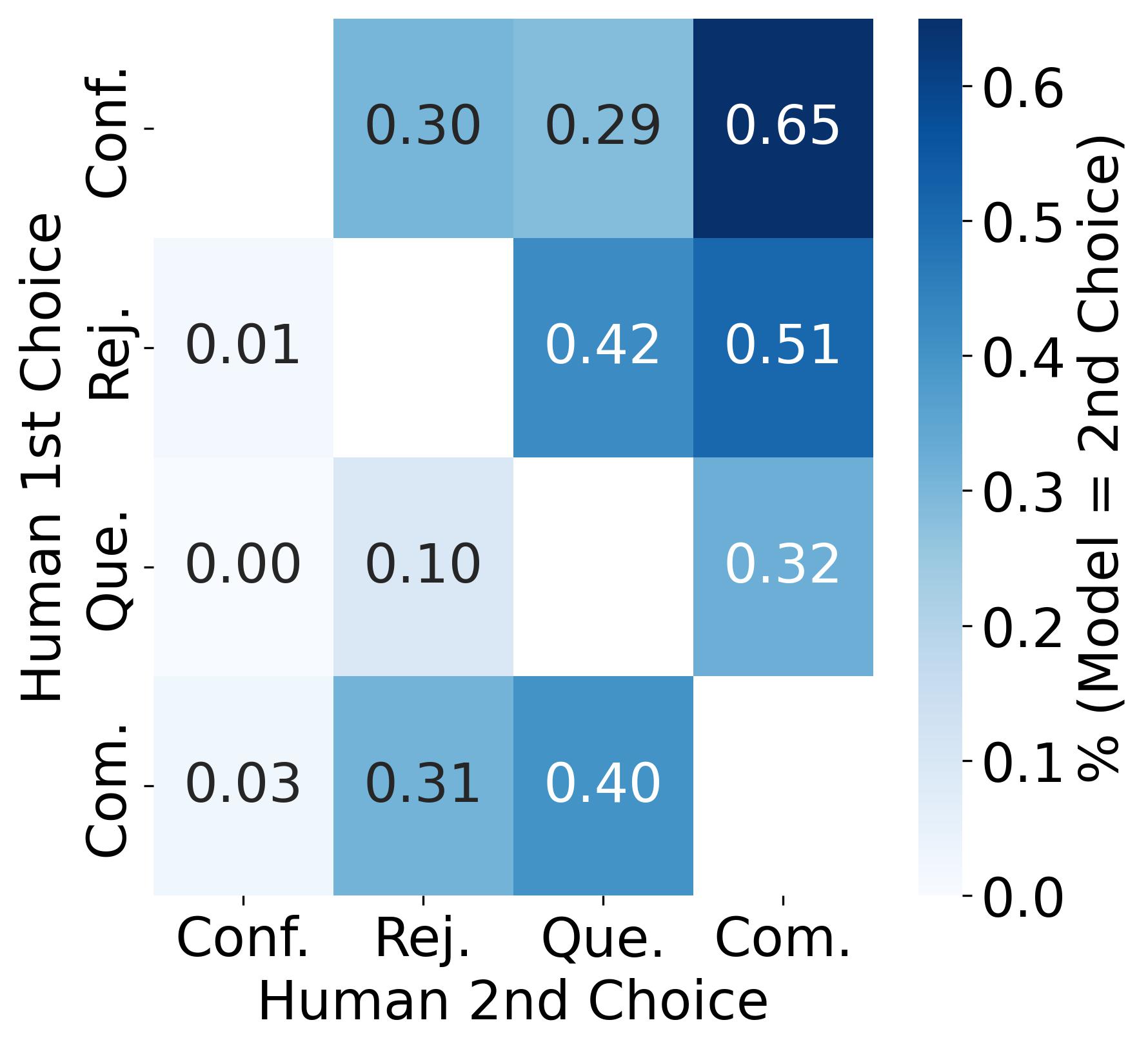}
\caption{Baseline Zero-Shot ICL (2nd Choice)}\label{fig:baseline-2}
\end{subfigure}
\hfill
\begin{subfigure}{0.2\textwidth}
\includegraphics[width=1\linewidth]{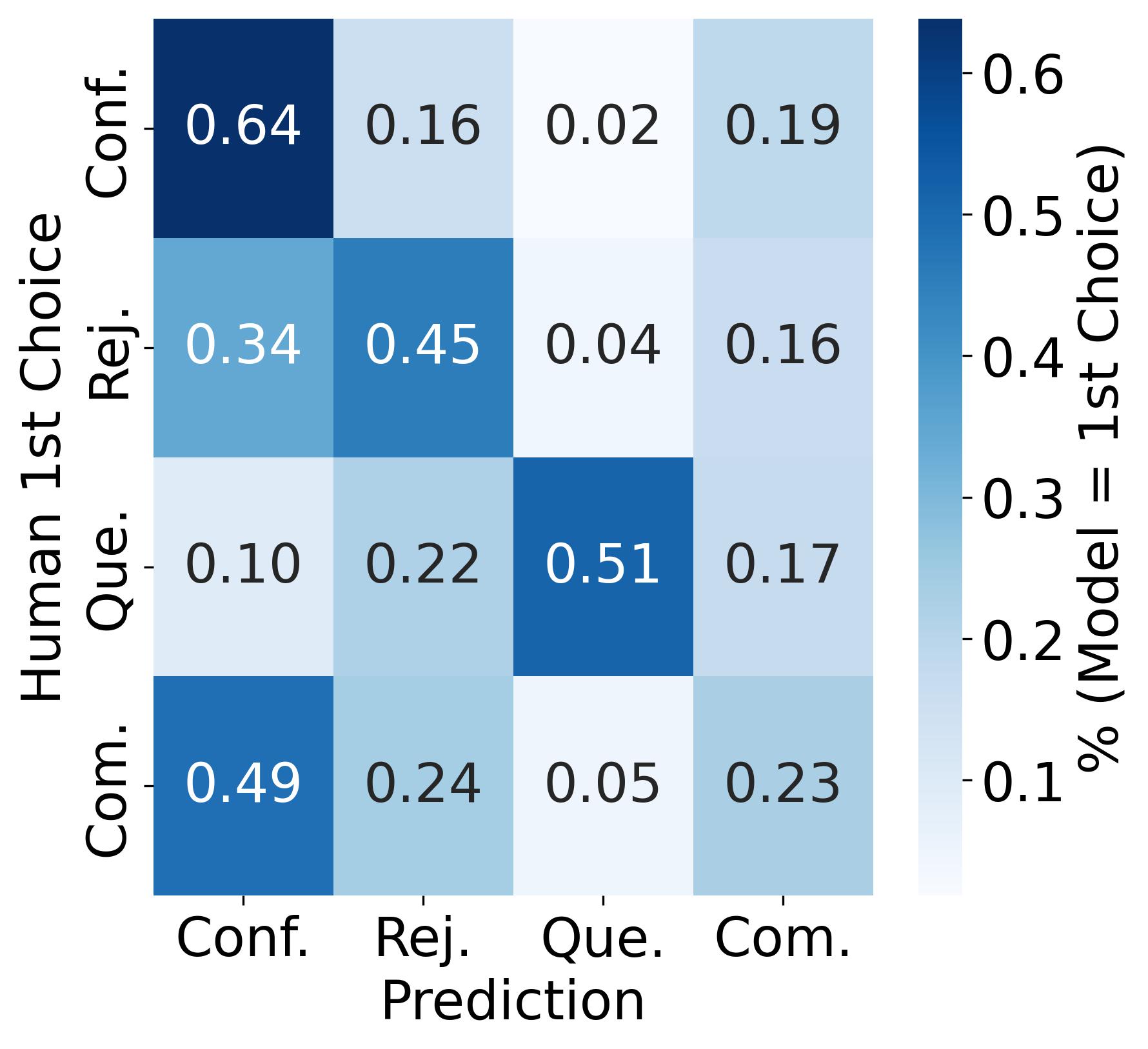}
\caption{XLM-R Train-Synthetic (1st Choice)}
\label{fig:synth-1}
\end{subfigure}
\hfill
\begin{subfigure}{0.2\textwidth}
\includegraphics[width=1\linewidth]{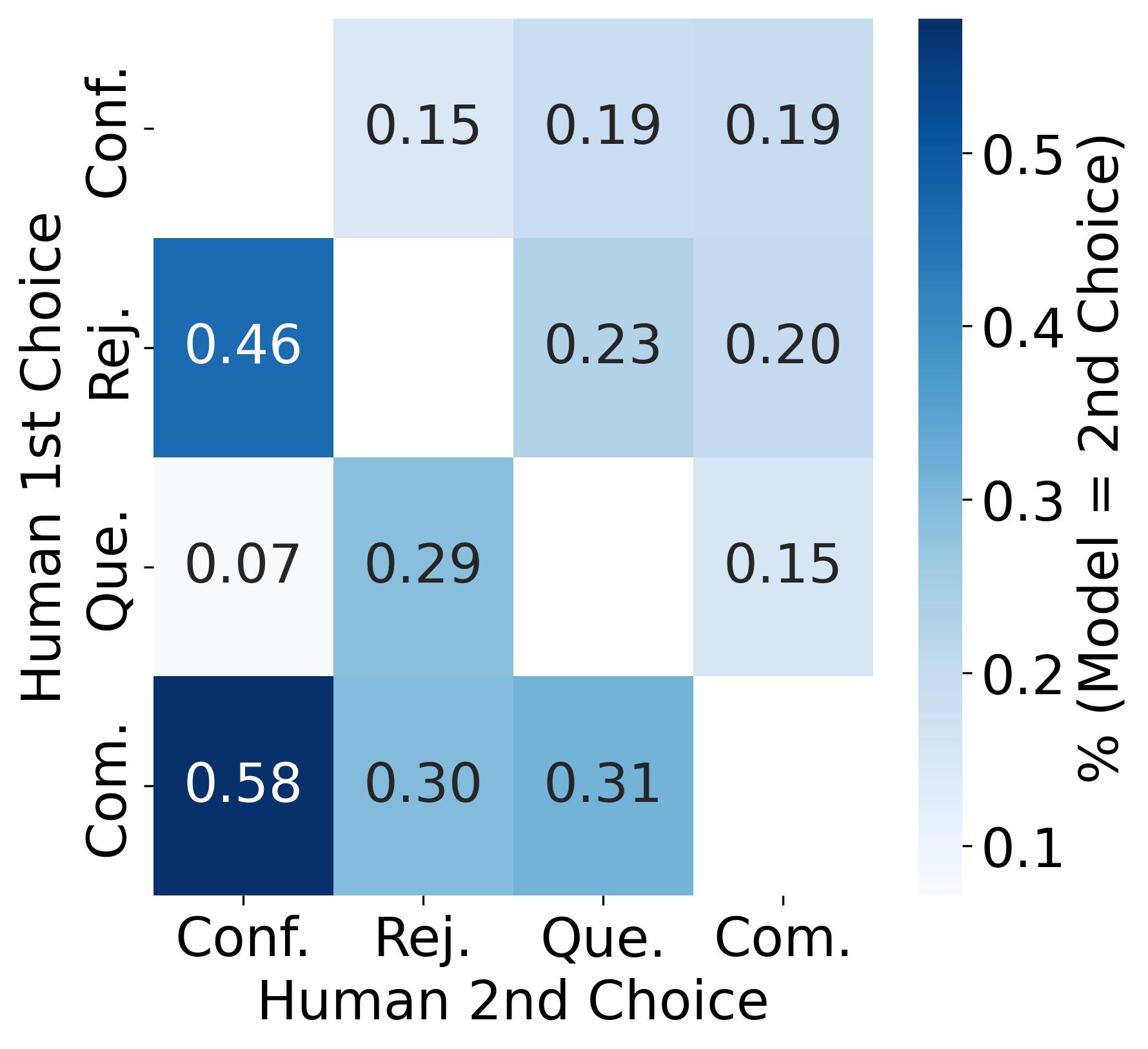}
\caption{XLM-R Train-Synthetic (2nd Choice)}
\label{fig:synth-2}
\end{subfigure}
\caption{Confusion matrices of (1) baseline zero-shot ICL performance with Deepseek; and (2) XLM-R performance when fine-tuned with sythetic data generated by Deepseek. \textit{In Sub-figures (a) and (c)}, each entry $(i, j)$ in row $i$ column $j$ represents the proportion of tweets (1st choice stance = $i$) that is classified as stance $j$ by the model. \textit{In Sub-figures (b) and (d)}, each entry $(i, j)$ denotes the proportion of tweets (1st choice stance = $i$, 2nd choice stance = $j$) that is classified as stance $j$ by the model.}
\label{fig:cm_second}
\end{figure*}

\paragraph{Can ICL performance predict the usefulness of synthetic data?} To investigate whether LLMs with stronger baseline ICL performance result in more effective synthetic data, we analyse the relationship between LLM's ICL performance and the performance of MLMs fine-tuned on synthetic data generated by this LLM.

Within each LLM across languages, we observe that \textbf{stronger ICL performance in a language tends to indicate more effective synthetic data generation} that leads to higher fine-tuning performance (as shown in Figure \ref{fig:ICL_syn}). Notably, \textbf{poor ICL performance does not necessarily result in poor fine-tuning outcomes}, which is consistent across all the five LLMs we study. For example, although Qwen’s baseline ICL on German is the lowest among all the languages in SCRum-9 ($wF2 = 0.21$ as shown in Figure \ref{fig:ICL_syn}), fine-tuning XLM-R on its synthetic data yields a $wF2$ score as 0.43, higher than other languages (e.g., Portuguese, Russian, French and Polish) where Qwen achieves better ICL performance but produces less effective synthetic data.

Furthermore, we find that the effectiveness of synthetic data also depends on the choice of MLM. For instance, Qwen’s synthetic data leads to the highest average performance on XLM-R, while Gemma’s generated data performs best on XLM-T.

Overall, practitioners could consider generating multilingual synthetic data for rumour stance classification in relatively high-resource languages with LLMs that exhibit strong ICL performance. \textbf{Fine-tuning MLMs on such data has the potential to match, and in some cases even exceed, the original ICL performance}.

\subsection{Model Prediction vs. Human Uncertainty} 

For ambiguous cases when human annotators provide a second-choice label, if the model predictions differ from the aggregated first-choice label but match the aggregated second-choice label, we assume the model still captures a plausible interpretation of the instance that aligns with human uncertainty. To capture this, we re-calculate our evaluation metrics by considering a prediction correct if it matches either the first-choice or second-choice label. 

We observe a substantial increase in $wF2$ scores across all the models, approaches and languages (full results see the Appendix). For example, $wF2$ score improves from 0.39 to 0.5 on Polish with Mistral in baseline zero-shot setting, when considering the second-choice human label. It suggests that, \textbf{in those ambiguous cases, the models are capable of generating predictions that reflect plausible alternative interpretations consistent with human uncertainty, not merely incorrect in a strict sense.}

As discussed before (also see Figure \ref{fig:first_second_label_cm}), most ambiguous cases involve annotator hesitation between \textit{confirming} and \textit{commenting} or between \textit{rejecting} and \textit{commenting}. We find that models handle these cases differently. Generally, MLMs trained with synthetic data exhibits different patterns than the other approaches we study in this paper. We present two examples related to Deepseek in Figure \ref{fig:cm_second}. The results suggest that in those ambiguous cases, Deepseek in baseline zero-shot ICL setting tends to predict \textit{commenting} (Figure \ref{fig:baseline-2}), whereas XLM-R fine-tuned with Deepseek-generated synthetic data more often predicts \textit{confirming} or \textit{rejecting} (Figure \ref{fig:synth-2}). This pattern is also reflected in their overall performance (Figure \ref{fig:baseline-1} and \ref{fig:synth-1}), where the former tends to over-predict \textit{commenting} and the latter tends to over-predict \textit{confirming} or \textit{rejecting}.

\section{Conclusion}

We introduce SCRum-9, the largest multilingual rumour stance classification benchmark to date, covering 7,516 instances in nine languages. The dataset incorporates annotator uncertainty through a second-choice label design, also enabling studies of uncertainty and subjectivity in stance classification. Our experiments reveal substantial performance disparities across languages in ICL with LLMs, while outperforming or matching MLMs fine-tuned with English or machine-translated multilingual data. However, MLMs fine-tuned on synthetic multilingual data generated by LLMs yields competitive results, in some cases surpassing ICL with the same LLM, while being more computationally efficient. We believe our work will serve as a valuable resource for advancing multilingual stance classification and online rumour analysis.

\section*{Limitations}
\label{sec:limitations}

While we go beyond previous work by assigning multiple annotators per example, budget and annotator availability limited us to only two or three annotators. This is less than ideal for obtaining a representative sample, so future work might want to obtain annotations from additional annotators. Nevertheless, we highlight the importance of our novel dataset, given the complexity of the task.

Our model evaluation did not take full advantage of the label aggregation methods, to our knowledge, there are no accepted methods for prompting LLMs to predict a distribution over labels. Future work ought to determine the extent to which LLMs can predict such distributions, and evaluate them against the various label aggregations described above. Also, although we analyse the effectiveness of LLM-generated synthetic data, we did not perform data analysis or human analysis to evaluate the multilingual data quality (e.g., lexical diversity or whether the generated replies truly express the stance required in the prompt) mainly due to lack of native speakers in the nine target languages. Although our experimental results have demonstrate their effectiveness, future work could explore how it correlates with the characteristics of the generated multilingual data.

While our evaluation focused on stance classification, because SCRum-9 links each source tweet to fact-checked claims, it is also possible to perform claim verification, which we did not evaluate here. Future work is required to build and evaluate claim verification models on our dataset.

Finally, although we make efforts to cover nine languages, the languages in this study are relatively high-resource languages, and future work should consider establishing datasets for low-resource languages, and benefiting more under-represented communities.

\section*{Acknowledgments}

This work is funded by EMIF managed by the Calouste Gulbenkian Foundation\footnote{The sole responsibility for any content supported by the European Media and Information Fund lies with the author(s) and it may not necessarily reflect the positions of the EMIF and the Fund Partners, the Calouste Gulbenkian Foundation and the European University Institute.} under the "Supporting Research into Media, Disinformation and Information Literacy Across Europe" call (ExU -- project number: 291191)\footnote{\url{exuproject.sites.sheffield.ac.uk}}. Yue Li is supported by a Sheffield–China Scholarships Council PhD Scholarship.

\bibliography{anthology,custom}

\begin{thebibliography}{40}
\providecommand{\natexlab}[1]{#1}

\bibitem[{Agerri et~al.(2021)Agerri, Centeno, Espinosa, de~Landa, and Rodrigo}]{agerri2021vaxxstance}
Agerri, R.; Centeno, R.; Espinosa, M.; de~Landa, J.~F.; and Rodrigo, A. 2021.
\newblock Vaxxstance@ iberlef 2021: Overview of the task on going beyond text in cross-lingual stance detection.
\newblock \emph{Procesamiento del lenguaje natural}, 67: 173--181.

\bibitem[{Barbieri, Espinosa~Anke, and Camacho-Collados(2022)}]{barbieri-etal-2022-xlm}
Barbieri, F.; Espinosa~Anke, L.; and Camacho-Collados, J. 2022.
\newblock {XLM}-{T}: Multilingual Language Models in {T}witter for Sentiment Analysis and Beyond.
\newblock In Calzolari, N.; B{\'e}chet, F.; Blache, P.; Choukri, K.; Cieri, C.; Declerck, T.; Goggi, S.; Isahara, H.; Maegaard, B.; Mariani, J.; Mazo, H.; Odijk, J.; and Piperidis, S., eds., \emph{Proceedings of the Thirteenth Language Resources and Evaluation Conference}, 258--266. Marseille, France: European Language Resources Association.

\bibitem[{Barriere, Jacquet, and Hemamou(2022)}]{barriere2022cofe}
Barriere, V.; Jacquet, G.~G.; and Hemamou, L. 2022.
\newblock CoFE: A new dataset of intra-multilingual multi-target stance classification from an online European participatory democracy platform.
\newblock In \emph{Proceedings of the 2nd Conference of the Asia-Pacific Chapter of the Association for Computational Linguistics and the 12th International Joint Conference on Natural Language Processing (Volume 2: Short Papers)}, 418--422.

\bibitem[{Conneau et~al.(2020)Conneau, Khandelwal, Goyal, Chaudhary, Wenzek, Guzm{\'a}n, Grave, Ott, Zettlemoyer, and Stoyanov}]{conneau-etal-2020-unsupervised}
Conneau, A.; Khandelwal, K.; Goyal, N.; Chaudhary, V.; Wenzek, G.; Guzm{\'a}n, F.; Grave, E.; Ott, M.; Zettlemoyer, L.; and Stoyanov, V. 2020.
\newblock Unsupervised Cross-lingual Representation Learning at Scale.
\newblock In Jurafsky, D.; Chai, J.; Schluter, N.; and Tetreault, J., eds., \emph{Proceedings of the 58th Annual Meeting of the Association for Computational Linguistics}, 8440--8451. Online: Association for Computational Linguistics.

\bibitem[{Cook et~al.(2025)Cook, Grimshaw, Wu, Dillon, Hicks, Jones, Smith, Szert, and Song}]{cook-etal-2025-efficient}
Cook, O.; Grimshaw, C.; Wu, B.~P.; Dillon, S.; Hicks, J.; Jones, L.; Smith, T.; Szert, M.; and Song, X. 2025.
\newblock Efficient Annotator Reliability Assessment and Sample Weighting for Knowledge-Based Misinformation Detection on Social Media.
\newblock In Chiruzzo, L.; Ritter, A.; and Wang, L., eds., \emph{Findings of the Association for Computational Linguistics: NAACL 2025}, 3348--3358. Albuquerque, New Mexico: Association for Computational Linguistics.
\newblock ISBN 979-8-89176-195-7.

\bibitem[{Dawid and Skene(1979)}]{dawid1979maximum}
Dawid, A.~P.; and Skene, A.~M. 1979.
\newblock {Maximum Likelihood Estimation of Observer Error-rates using the EM Algorithm}.
\newblock \emph{Journal of the Royal Statistical Society: Series C (Applied Statistics)}, 28(1): 20--28.

\bibitem[{Derczynski et~al.(2017)Derczynski, Bontcheva, Liakata, Procter, Wong Sak~Hoi, and Zubiaga}]{derczynski-etal-2017-semeval}
Derczynski, L.; Bontcheva, K.; Liakata, M.; Procter, R.; Wong Sak~Hoi, G.; and Zubiaga, A. 2017.
\newblock {S}em{E}val-2017 Task 8: {R}umour{E}val: Determining rumour veracity and support for rumours.
\newblock In Bethard, S.; Carpuat, M.; Apidianaki, M.; Mohammad, S.~M.; Cer, D.; and Jurgens, D., eds., \emph{Proceedings of the 11th International Workshop on Semantic Evaluation ({S}em{E}val-2017)}, 69--76. Vancouver, Canada: Association for Computational Linguistics.

\bibitem[{Dumitrache et~al.(2018)Dumitrache, Inel, Aroyo, Timmermans, and Welty}]{dumitrache2018crowdtruth}
Dumitrache, A.; Inel, O.; Aroyo, L.; Timmermans, B.; and Welty, C. 2018.
\newblock CrowdTruth 2.0: Quality metrics for crowdsourcing with disagreement.
\newblock \emph{arXiv preprint arXiv:1808.06080}.

\bibitem[{Espa{\~n}a-Bonet(2023)}]{espana-bonet-2023-multilingual}
Espa{\~n}a-Bonet, C. 2023.
\newblock Multilingual Coarse Political Stance Classification of Media. The Editorial Line of a {C}hat{GPT} and Bard Newspaper.
\newblock In Bouamor, H.; Pino, J.; and Bali, K., eds., \emph{Findings of the Association for Computational Linguistics: EMNLP 2023}, 11757--11777. Singapore: Association for Computational Linguistics.

\bibitem[{Garc{\'i}a~Lozano et~al.(2017)Garc{\'i}a~Lozano, Lilja, Tj{\"o}rnhammar, and Karasalo}]{garcia-lozano-etal-2017-mama}
Garc{\'i}a~Lozano, M.; Lilja, H.; Tj{\"o}rnhammar, E.; and Karasalo, M. 2017.
\newblock Mama Edha at {S}em{E}val-2017 Task 8: Stance Classification with {CNN} and Rules.
\newblock In Bethard, S.; Carpuat, M.; Apidianaki, M.; Mohammad, S.~M.; Cer, D.; and Jurgens, D., eds., \emph{Proceedings of the 11th International Workshop on Semantic Evaluation ({S}em{E}val-2017)}, 481--485. Vancouver, Canada: Association for Computational Linguistics.

\bibitem[{Gorrell et~al.(2019{\natexlab{a}})Gorrell, Kochkina, Liakata, Aker, Zubiaga, Bontcheva, and Derczynski}]{gorrell2019semeval}
Gorrell, G.; Kochkina, E.; Liakata, M.; Aker, A.; Zubiaga, A.; Bontcheva, K.; and Derczynski, L. 2019{\natexlab{a}}.
\newblock Semeval-2019 task 7: Rumoureval 2019: Determining rumour veracity and support for rumours.
\newblock In \emph{Proceedings of the 13th International Workshop on Semantic Evaluation: NAACL HLT 2019}, 845--854. Association for Computational Linguistics.

\bibitem[{Gorrell et~al.(2019{\natexlab{b}})Gorrell, Kochkina, Liakata, Aker, Zubiaga, Bontcheva, and Derczynski}]{gorrell-etal-2019-semeval}
Gorrell, G.; Kochkina, E.; Liakata, M.; Aker, A.; Zubiaga, A.; Bontcheva, K.; and Derczynski, L. 2019{\natexlab{b}}.
\newblock {S}em{E}val-2019 Task 7: {R}umour{E}val, Determining Rumour Veracity and Support for Rumours.
\newblock In May, J.; Shutova, E.; Herbelot, A.; Zhu, X.; Apidianaki, M.; and Mohammad, S.~M., eds., \emph{Proceedings of the 13th International Workshop on Semantic Evaluation}, 845--854. Minneapolis, Minnesota, USA: Association for Computational Linguistics.

\bibitem[{Grattafiori et~al.(2024)Grattafiori, Dubey, Jauhri, Pandey, Kadian, Al-Dahle, Letman, Mathur, Schelten, Vaughan et~al.}]{grattafiori2024llama}
Grattafiori, A.; Dubey, A.; Jauhri, A.; Pandey, A.; Kadian, A.; Al-Dahle, A.; Letman, A.; Mathur, A.; Schelten, A.; Vaughan, A.; et~al. 2024.
\newblock The llama 3 herd of models.
\newblock \emph{arXiv preprint arXiv:2407.21783}.

\bibitem[{Grootendorst(2022)}]{grootendorst2022bertopicneuraltopicmodeling}
Grootendorst, M. 2022.
\newblock {BERTopic: Neural topic modeling with a class-based TF-IDF procedure}.
\newblock arXiv:2203.05794.

\bibitem[{Hamdi et~al.(2021)Hamdi, Linhares~Pontes, Boros, Nguyen, Hackl, Moreno, and Doucet}]{hamdi2021multilingual}
Hamdi, A.; Linhares~Pontes, E.; Boros, E.; Nguyen, T. T.~H.; Hackl, G.; Moreno, J.~G.; and Doucet, A. 2021.
\newblock A multilingual dataset for named entity recognition, entity linking and stance detection in historical newspapers.
\newblock In \emph{Proceedings of the 44th International ACM SIGIR Conference on Research and Development in Information Retrieval}, 2328--2334.

\bibitem[{Hardalov et~al.(2022{\natexlab{a}})Hardalov, Arora, Nakov, and Augenstein}]{hardalov2022few}
Hardalov, M.; Arora, A.; Nakov, P.; and Augenstein, I. 2022{\natexlab{a}}.
\newblock Few-shot cross-lingual stance detection with sentiment-based pre-training.
\newblock In \emph{Proceedings of the AAAI Conference on Artificial Intelligence}, volume~36, 10729--10737.

\bibitem[{Hardalov et~al.(2022{\natexlab{b}})Hardalov, Arora, Nakov, and Augenstein}]{hardalov-etal-2022-survey}
Hardalov, M.; Arora, A.; Nakov, P.; and Augenstein, I. 2022{\natexlab{b}}.
\newblock A Survey on Stance Detection for Mis- and Disinformation Identification.
\newblock In Carpuat, M.; de~Marneffe, M.-C.; and Meza~Ruiz, I.~V., eds., \emph{Findings of the Association for Computational Linguistics: NAACL 2022}, 1259--1277. Seattle, United States: Association for Computational Linguistics.

\bibitem[{Hardalov et~al.(2022{\natexlab{c}})Hardalov, Arora, Nakov, and Augenstein}]{hardalov2022survey}
Hardalov, M.; Arora, A.; Nakov, P.; and Augenstein, I. 2022{\natexlab{c}}.
\newblock A Survey on Stance Detection for Mis-and Disinformation Identification.
\newblock In \emph{Findings of the Association for Computational Linguistics: NAACL 2022}, 1259--1277.

\bibitem[{Kochkina and Liakata(2020)}]{kochkina-liakata-2020-estimating}
Kochkina, E.; and Liakata, M. 2020.
\newblock Estimating predictive uncertainty for rumour verification models.
\newblock In Jurafsky, D.; Chai, J.; Schluter, N.; and Tetreault, J., eds., \emph{Proceedings of the 58th Annual Meeting of the Association for Computational Linguistics}, 6964--6981. Online: Association for Computational Linguistics.

\bibitem[{Lavrouk et~al.(2024)Lavrouk, Ligon, Zheng, Naous, Xu, and Ritter}]{lavrouk-etal-2024-stanceosaurus}
Lavrouk, A.; Ligon, I.; Zheng, J.; Naous, T.; Xu, W.; and Ritter, A. 2024.
\newblock Stanceosaurus 2.0 - Classifying Stance Towards {R}ussian and {S}panish Misinformation.
\newblock In van~der Goot, R.; Bak, J.; M{\"u}ller-Eberstein, M.; Xu, W.; Ritter, A.; and Baldwin, T., eds., \emph{Proceedings of the Ninth Workshop on Noisy and User-generated Text (W-NUT 2024)}, 31--43. San {\.{G}}iljan, Malta: Association for Computational Linguistics.

\bibitem[{Le~Scao et~al.(2023)Le~Scao, Fan, Akiki, Pavlick, Ili{\'c}, Hesslow, Castagn{\'e}, Luccioni, Yvon, Gall{\'e} et~al.}]{le2023bloom}
Le~Scao, T.; Fan, A.; Akiki, C.; Pavlick, E.; Ili{\'c}, S.; Hesslow, D.; Castagn{\'e}, R.; Luccioni, A.~S.; Yvon, F.; Gall{\'e}, M.; et~al. 2023.
\newblock Bloom: A 176b-parameter open-access multilingual language model.

\bibitem[{Li, Zhang, and Si(2019)}]{li-etal-2019-eventai}
Li, Q.; Zhang, Q.; and Si, L. 2019.
\newblock event{AI} at {S}em{E}val-2019 Task 7: Rumor Detection on Social Media by Exploiting Content, User Credibility and Propagation Information.
\newblock In May, J.; Shutova, E.; Herbelot, A.; Zhu, X.; Apidianaki, M.; and Mohammad, S.~M., eds., \emph{Proceedings of the 13th International Workshop on Semantic Evaluation}, 855--859. Minneapolis, Minnesota, USA: Association for Computational Linguistics.

\bibitem[{Li and Scarton(2024)}]{li-scarton-2024-identify}
Li, Y.; and Scarton, C. 2024.
\newblock Can We Identify Stance without Target Arguments? A Study for Rumour Stance Classification.
\newblock In Calzolari, N.; Kan, M.-Y.; Hoste, V.; Lenci, A.; Sakti, S.; and Xue, N., eds., \emph{Proceedings of the 2024 Joint International Conference on Computational Linguistics, Language Resources and Evaluation (LREC-COLING 2024)}, 2844--2851. Torino, Italia: ELRA and ICCL.

\bibitem[{Li, Zhao, and Scarton(2025)}]{li2025s}
Li, Y.; Zhao, Z.; and Scarton, C. 2025.
\newblock It's All About In-Context Learning! Teaching Extremely Low-Resource Languages to LLMs.
\newblock \emph{arXiv preprint arXiv:2508.19089}.

\bibitem[{Mu et~al.(2023)Mu, Jin, Grimshaw, Scarton, Bontcheva, and Song}]{mu2023vaxxhesitancy}
Mu, Y.; Jin, M.; Grimshaw, C.; Scarton, C.; Bontcheva, K.; and Song, X. 2023.
\newblock Vaxxhesitancy: A dataset for studying hesitancy towards covid-19 vaccination on twitter.
\newblock In \emph{Proceedings of the International AAAI Conference on Web and Social Media}, volume~17, 1052--1062.

\bibitem[{Razumovskaia, Vulić, and Korhonen(2025)}]{RazumovskaiaAnalyzing}
Razumovskaia, E.; Vulić, I.; and Korhonen, A. 2025.
\newblock Analyzing and Adapting Large Language Models for Few-Shot Multilingual NLU: Are We There Yet?
\newblock \emph{Transactions of the Association for Computational Linguistics}, 13: 1096--1120.

\bibitem[{Scarton and Li(2021)}]{scarton2021cross}
Scarton, C.; and Li, Y. 2021.
\newblock Cross-lingual Rumour Stance Classification: a First Study with BERT and Machine Translation.
\newblock In \emph{TTO}, 50--59.

\bibitem[{Scarton, Silva, and Bontcheva(2020{\natexlab{a}})}]{scarton-etal-2020-measuring}
Scarton, C.; Silva, D.; and Bontcheva, K. 2020{\natexlab{a}}.
\newblock Measuring What Counts: The Case of Rumour Stance Classification.
\newblock In Wong, K.-F.; Knight, K.; and Wu, H., eds., \emph{Proceedings of the 1st Conference of the Asia-Pacific Chapter of the Association for Computational Linguistics and the 10th International Joint Conference on Natural Language Processing}, 925--932. Suzhou, China: Association for Computational Linguistics.

\bibitem[{Scarton, Silva, and Bontcheva(2020{\natexlab{b}})}]{scarton2020measuring}
Scarton, C.; Silva, D.; and Bontcheva, K. 2020{\natexlab{b}}.
\newblock Measuring What Counts: The Case of Rumour Stance Classification.
\newblock In \emph{Proceedings of the 1st Conference of the Asia-Pacific Chapter of the Association for Computational Linguistics and the 10th International Joint Conference on Natural Language Processing}, 925--932.

\bibitem[{Shu et~al.(2020)Shu, Mahudeswaran, Wang, Lee, and Liu}]{shu2020combating}
Shu, K.; Mahudeswaran, D.; Wang, S.; Lee, D.; and Liu, H. 2020.
\newblock Combating disinformation in a social media age.
\newblock \emph{Wiley Interdisciplinary Reviews: Data Mining and Knowledge Discovery}, 10(6): e1385.

\bibitem[{Vamvas and Sennrich(2020)}]{vamvas2020xstance}
Vamvas, J.; and Sennrich, R. 2020.
\newblock {X-Stance}: A Multilingual Multi-Target Dataset for Stance Detection.
\newblock In \emph{Proceedings of the 5th Swiss Text Analytics Conference (SwissText) \& 16th Conference on Natural Language Processing (KONVENS)}. Zurich, Switzerland.

\bibitem[{Wilby et~al.(2023)Wilby, Karmakharm, Roberts, Song, and Bontcheva}]{wilby-etal-2023-gate}
Wilby, D.; Karmakharm, T.; Roberts, I.; Song, X.; and Bontcheva, K. 2023.
\newblock {GATE} Teamware 2: An open-source tool for collaborative document classification annotation.
\newblock In Croce, D.; and Soldaini, L., eds., \emph{Proceedings of the 17th Conference of the European Chapter of the Association for Computational Linguistics: System Demonstrations}, 145--151. Dubrovnik, Croatia: Association for Computational Linguistics.

\bibitem[{Wu et~al.(2023{\natexlab{a}})Wu, Li, Mu, Scarton, Bontcheva, and Song}]{wu-etal-2023-dont}
Wu, B.; Li, Y.; Mu, Y.; Scarton, C.; Bontcheva, K.; and Song, X. 2023{\natexlab{a}}.
\newblock Don`t waste a single annotation: improving single-label classifiers through soft labels.
\newblock In Bouamor, H.; Pino, J.; and Bali, K., eds., \emph{Findings of the Association for Computational Linguistics: EMNLP 2023}, 5347--5355. Singapore: Association for Computational Linguistics.

\bibitem[{Wu et~al.(2023{\natexlab{b}})Wu, Li, Mu, Scarton, Bontcheva, and Song}]{wu2023don}
Wu, B.; Li, Y.; Mu, Y.; Scarton, C.; Bontcheva, K.; and Song, X. 2023{\natexlab{b}}.
\newblock Don’t waste a single annotation: improving single-label classifiers through soft labels.
\newblock In \emph{Findings of the Association for Computational Linguistics: EMNLP 2023}, 5347--5355.

\bibitem[{Zhang, Yang, and Mao(2023)}]{zhang2023cross}
Zhang, R.; Yang, H.; and Mao, W. 2023.
\newblock Cross-lingual cross-target stance detection with dual knowledge distillation framework.
\newblock In \emph{Proceedings of the 2023 Conference on Empirical Methods in Natural Language Processing}, 10804--10819.

\bibitem[{Zhang et~al.(2023)Zhang, Li, Hauer, Shi, and Kondrak}]{zhang-etal-2023-dont}
Zhang, X.; Li, S.; Hauer, B.; Shi, N.; and Kondrak, G. 2023.
\newblock Don`t Trust {C}hat{GPT} when your Question is not in {E}nglish: A Study of Multilingual Abilities and Types of {LLM}s.
\newblock In Bouamor, H.; Pino, J.; and Bali, K., eds., \emph{Proceedings of the 2023 Conference on Empirical Methods in Natural Language Processing}, 7915--7927. Singapore: Association for Computational Linguistics.

\bibitem[{Zheng et~al.(2022)Zheng, Baheti, Naous, Xu, and Ritter}]{zheng-etal-2022-stanceosaurus}
Zheng, J.; Baheti, A.; Naous, T.; Xu, W.; and Ritter, A. 2022.
\newblock Stanceosaurus: Classifying Stance Towards Multicultural Misinformation.
\newblock In Goldberg, Y.; Kozareva, Z.; and Zhang, Y., eds., \emph{Proceedings of the 2022 Conference on Empirical Methods in Natural Language Processing}, 2132--2151. Abu Dhabi, United Arab Emirates: Association for Computational Linguistics.

\bibitem[{Zotova et~al.(2020)Zotova, Agerri, Nu{\~n}ez, and Rigau}]{zotova-etal-2020-multilingual}
Zotova, E.; Agerri, R.; Nu{\~n}ez, M.; and Rigau, G. 2020.
\newblock Multilingual Stance Detection in Tweets: The {C}atalonia Independence Corpus.
\newblock In Calzolari, N.; B{\'e}chet, F.; Blache, P.; Choukri, K.; Cieri, C.; Declerck, T.; Goggi, S.; Isahara, H.; Maegaard, B.; Mariani, J.; Mazo, H.; Moreno, A.; Odijk, J.; and Piperidis, S., eds., \emph{Proceedings of the Twelfth Language Resources and Evaluation Conference}, 1368--1375. Marseille, France: European Language Resources Association.
\newblock ISBN 979-10-95546-34-4.

\bibitem[{Zubiaga et~al.(2018)Zubiaga, Aker, Bontcheva, Liakata, and Procter}]{zubiaga-etal-2018-survey}
Zubiaga, A.; Aker, A.; Bontcheva, K.; Liakata, M.; and Procter, R. 2018.
\newblock Detection and Resolution of Rumours in Social Media: A Survey.
\newblock \emph{ACM Comput. Surv.}, 51(2).

\bibitem[{Zubiaga et~al.(2016)Zubiaga, Liakata, Procter, Wong Sak~Hoi, and Tolmie}]{zubiaga2016analysing}
Zubiaga, A.; Liakata, M.; Procter, R.; Wong Sak~Hoi, G.; and Tolmie, P. 2016.
\newblock Analysing how people orient to and spread rumours in social media by looking at conversational threads.
\newblock \emph{PloS one}, 11(3): e0150989.

\end{thebibliography}

\section*{Ethical Considerations}
\label{sec:ethics}

We anonymised the tweets by removing user mentions. This was done primarily for anonymity during annotation. However, because the tweets are indexed in our dataset by tweet ID, it is simple to deanonymise them by looking them up on X directly or using the API. Keying by the tweet IDs is necessary for our dataset to be used by other researchers, as the X developer agreement prohibits redistribution of tweet texts, instead requiring users to use the X API to hydrate tweets keyed by their ID. The dataset will be released with a CC BY-NC-SA 4.0 license.\footnote{\url{https://creativecommons.org/licenses/by-nc-sa/4.0/deed.en}}

We received ethics approval by our institution (omitted due to double-blind constraints) to conduct this research. We consider social media data as a type of personal data where consent is not feasible to obtain. Nevertheless, following legislation, we are allowed to collect data which will result in research of public interest. Since the collected tweets are flagged by fact-checkers as potential source of disinformation, their collection is justified. In addition, we also received ethical approval for the annotation experiments, complying with the best practices for information sheet and consent forms creation as well as data anonymisation prior to annotation. 

All data is securely stored in our encrypted servers. The data shared with annotators is done via annotation tool, that also  stores  all data in an encrypted servers.  Annotators  data  (e.g.  e-mail) are stored separately from the annotations and researchers only have  access  to  annotators’ IDs and annotators (i.e.  no  access to personal information is  available). The dataset is made available for research without any identification to the annotators. 

Finally, we advise all users of our dataset that because the tweets were sourced from X they contain potentially offensive content. 

\appendix

\section{Data Collection and Filtering} \label{app:datacollection}

We provide URLs of each fact-checking source used during data collection in \cref{tab:data_collection_sources}. Additional details and scripts used to obtain data from DBKF and the fact-checking websites will be made available with the publicly available source code. The statistics of the source tweets, replies, and fact-checked claims that were collected, filtered, and annotated can be found in Table \ref{tab:collected_tweet_stats}.

\begin{table}[h!]
\centering
\begin{tabular}{c|l}
\toprule
CS & \url{dbkf.ontotext.com};  \url{napravoumiru.afp.com}   \\
& \url{demagog.cz}   \\
DE & \url{dbkf.ontotext.com}      \\
EN & \url{factcheck.afp.com}; \url{dbkf.ontotext.com} \\
ES & \url{factual.afp.com}; \url{dbkf.ontotext.com}       \\
FR & \url{factuel.afp.com}; \url{dbkf.ontotext.com}  \\
PL & \url{sprawdzam.afp.com}; \url{demagog.org.pl}     \\
& \url{dbkf.ontotext.com}      \\
PT & \url{checamos.afp.com}; \url{dbkf.ontotext.com}  \\
RU & \url{dbkf.ontotext.com}      \\
\bottomrule
\end{tabular}
\caption{Source fact-check websites used for sourcing the X links.}
\label{tab:data_collection_sources}
\end{table}

\begin{table*}[t]
    \centering
    \small
    \begin{tabular}{c|rrr|rrr|rrr}
        \toprule
              & \multicolumn{3}{c|}{Collected} & \multicolumn{3}{c|}{Filtered}  &\multicolumn{3}{c}{Annotated} \\
         Lang &Sources& Replies   &  Claims    &  Sources& Replies    & Claims & Sources&  Replies & Claims   \\
        \midrule
          CS  & 89    & 918       &  128       &  24     & 780        &   14   &   24   &   780    &   14     \\ 
          DE  & 358   & 4,548     &  528       &  49     & 1,500      &   56   &   24   &   705    &   19     \\
          EN  & 15,734& 215,152   &  19,529    &  1,402  & 1,500      &  1,565 &  1,026 &   1,100  &  1,169   \\
          ES  & 7,894 & 70,333    &  11,518    &  491    & 1,500      &    530 &  414   &   1,260  &  419     \\
          FR  & 1,477 & 20,159    &  2,864     &  224    & 1,500      &   286  &  109   &   700    &  122     \\
          HI  & 1,668 & 13,601    &  2,130     &  320    & 1,500      &   342  &  220   &   1,000  &  227     \\
          PL  &  258  & 4,105     &  404       &  42     & 1,500      &   45   &  23    &   700    &  224     \\
          PT  & 2,485 & 22,473    &  3,845     &  197    & 1,500      &   204  &  124   &   1,000  &  133     \\
          RU  &  59   &  831      &  95        &  20     &   495      &   31   &  19    &   450    &  29      \\
        \midrule
        All   & 30,022& 352,120   &  41,041    & 2,769   & 11,775     &  3,073 &  1,983 &   7,695  &  2,156   \\
        \bottomrule
    \end{tabular}
    \caption{Statistics of the source tweets, replies, and fact-checked claims that were collected, filtered, and annotated.}
    \label{tab:collected_tweet_stats}
\end{table*}

We use the \texttt{bertopic} Python library to perform the topic modelling.\footnote{\url{https://maartengr.github.io/BERTopic/index.html}} Specifically, we use KeyBertInspired as the representation model and a UMAP model 15 neighbours, 5 components, and cosine distance to perform dimensionality reduction. We then perform a manual review of the resulting topics, keeping only those that we deemed coherent, and merging topics that were closely related. Of the resulting 63 topics, the five most common are Russia-Ukraine (774 source tweets), COVID-19 (440), US Elections (301), Israel-Palestine (201), and Natural Disasters (133), all of which are represented by the six languages (DE, EN, ES, FR, PL, PT) for which we used topic model filtering. We also report the five most common topics per language in \cref{tab:topic_per_language}. 

\begin{table}[h]
    \centering
    \scalebox{0.9}{
    \begin{tabular}{c|p{0.85\linewidth}}
        \toprule
         DE & Russia-Ukraine, COVID-19, US Elections, Climate Change, Israel-Palestine  \\
         EN & Russia-Ukraine, COVID-19, US Elections, Israel-Palestine, Natural Disasters \\
         ES & Russia-Ukraine, COVID-19, Venezuela Dictatorship, US Elections, Natural Disasters \\
         FR & Russia-Ukraine, COVID-19, Climate Change, US Elections, Israel-Palestine \\
         PL & Russia-Ukraine, COVID-19, US tax legislation, The Kashmir Files Film, US Immigration \\
         PT & COVID-19, Brazil President Lula, Russia-Ukraine, US Elections, COVID-19 \\
        \bottomrule
    \end{tabular}
    }
    \caption{The top five most common topics per language.}
    \label{tab:topic_per_language}
\end{table}

\section{Annotation Guideline and Interface}
\label{sec:annotation_guidelines}

\begin{figure*}[h!]
    \centering
    \scalebox{0.8}{
    \includegraphics[width=\linewidth]{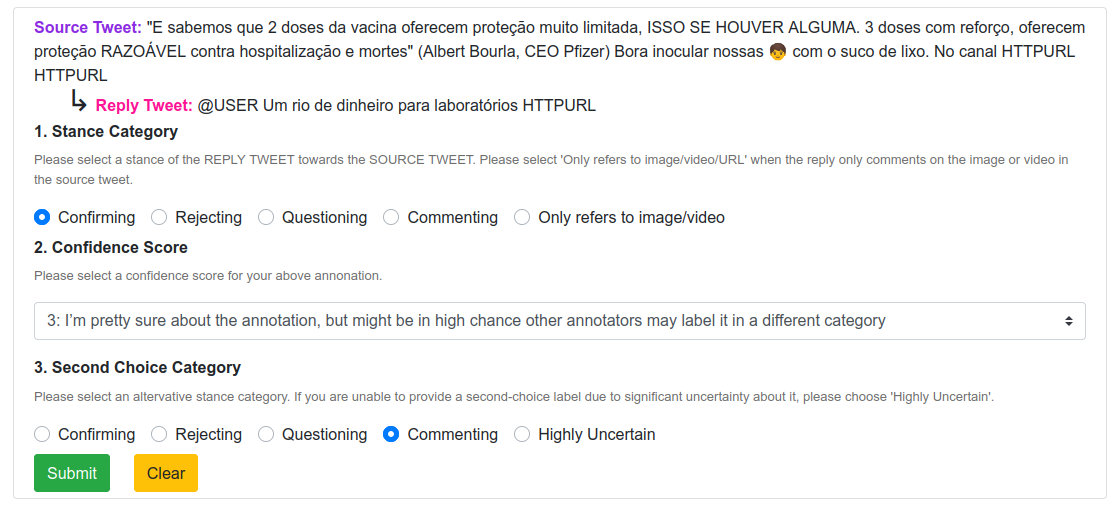}
    }
    \caption{An example of the GATE Teamware annotation interface. The Second Choice Category section only appears when the provided Confidence Score is 3 or lower. The tool also ensures that annotators do not choose the same label for the first and second choice.} 
    \label{fig:gate_screenshot}
\end{figure*}

A screenshot of the annotation interface is presented in Figure \ref{fig:gate_screenshot}. The full text of the annotation guidelines is reproduced below:

\begin{enumerate}

\item \textbf{Stance Annotation:} After reading the source and reply tweets thoroughly, you will need to decide the stance of the reply tweet towards the source tweet based on the following definitions. Please ONLY rely on the given tweets and refrain from using additional resources in this step.
\begin{itemize}
    \item Confirming: The author of the response supports the source tweet.
    \item Rejecting: The author of the response disagrees with the source tweet.
    \item Questioning: The author of the response asks for additional evidence in relation to the source tweet.
    \item Commenting: The author of the response makes their own comment without a clear stance towards the source tweet. It includes replies that are unrelated to the source tweet.
    \item Only refers to image/video: The author of the response makes the comment only/mainly referring to the image or video attached to the source tweet.
\end{itemize}

Note that sarcastically or humorously supporting the source tweet should be considered as rejecting, while sarcastically or humorously denying source tweet should be considered as confirming. Also, the appearance of question marks does NOT necessarily indicate the questioning stance. For example, rhetorical questions should not be coded as questioning, since the author does not expect an answer. 

\item \textbf{Confidence Rating:} Please indicate how confident you are about your annotation. The confidence scores range from 1 to 5 and hold the following meaning. You will need to indicate a second-choice stance label if your confidence score is lower than 3, including 3.

\begin{itemize}
    \item 5 - extremely confident about the annotation (I’m certain about the annotation without a doubt.)
    \item 4 - fairly confident about the annotation (I’m confident about the annotation, but might be in small chance other annotators may label it in a different category)
    \item 3 - pretty confident about the annotation (I’m pretty sure about the annotation, but might be in high chance other annotators may label it in a different category)
    \item 2 - not confident about the annotation (I’m not sure about the annotation, it seems it also belongs to other categories, but you can still include this instance as a “silver standard instance” in training)
    \item 1 - extremely unconfident about the annotation (I’m really unsure about the annotation. It may belong to another category as well, you may wish to discard this instance from the training.)
\end{itemize}

\item \textbf{Second-choice Label:} If your confidence score is lower than 3 including 3, please select an alternative category that the stance of the target tweet may also belong to, except for the following conditions: (1) \textit{Not Applicable}: Your confidence score is larger than 3; (2) \textit{Highly Uncertain}: You are unable to provide a second-choice label due to significant uncertainty about it. 

\end{enumerate}

\section{Second-Choice Label Analysis}

\begin{figure}[h!]
    \centering
    \scalebox{0.6}{
    \includegraphics[width=\linewidth]{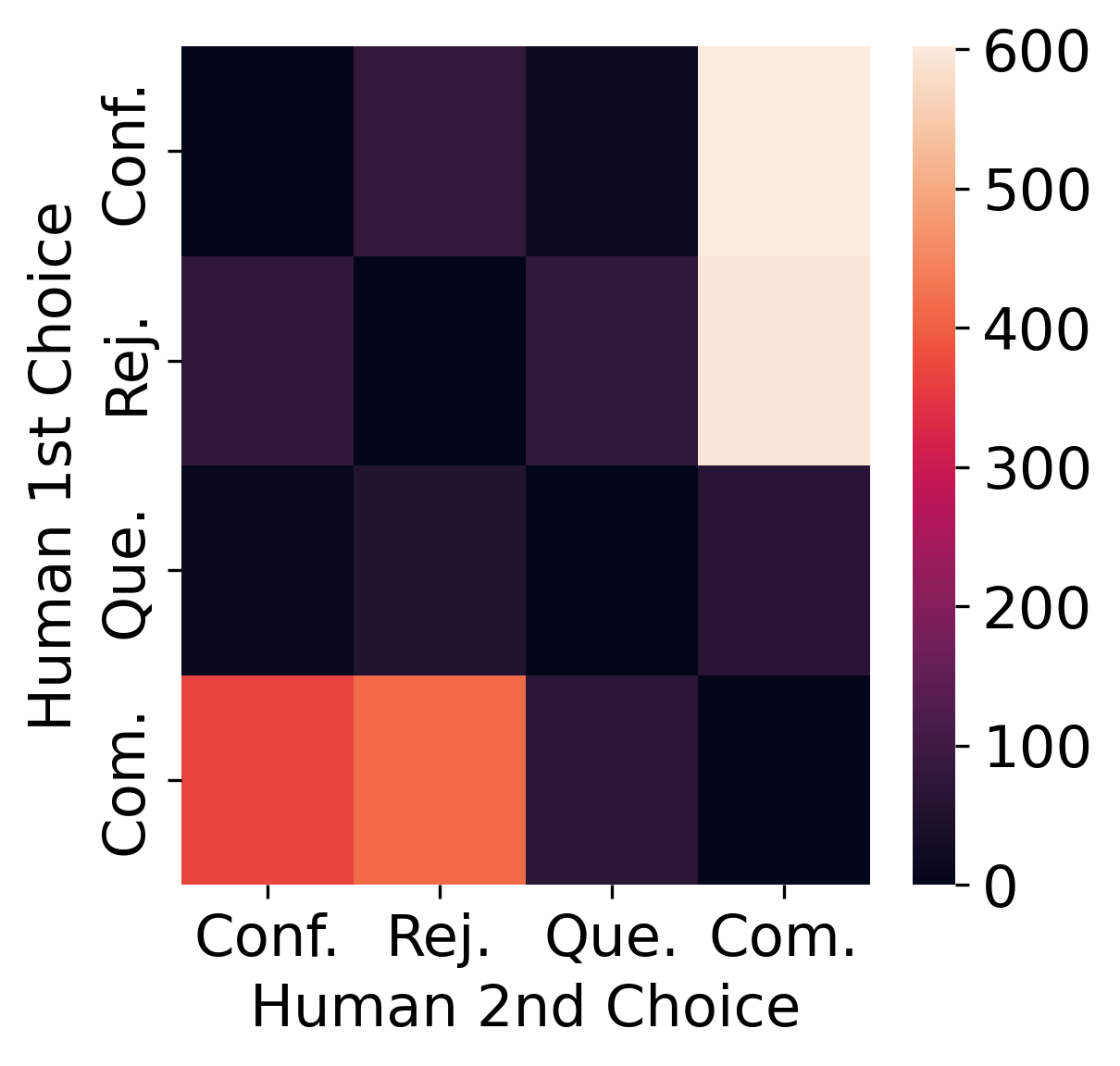}
    }
    \caption{Confusion matrix between annotators' first-choice and second-choice labels. Each entry $(i,j)$ in row $i$ column $j$ denotes the number of tweets whose first-choice label is stance $i$ and second-choice label is stance $j$. The labels are aggregated through majority-voting.}   \label{fig:first_second_label_cm}
\end{figure}

Figure \ref{fig:first_second_label_cm} presents the confusion matrix between annotators’ first-choice and second-choice labels, aggregated via majority voting. The results show that, in most ambiguous cases, annotators select \textit{confirming} or \textit{rejecting} as the first-choice label while considering \textit{commenting} as a possible alternative. The next most frequent pattern is the reverse: annotators select \textit{commenting} as the first-choice label but also consider \textit{confirming} or \textit{rejecting} as plausible options.

\section{Label Aggregation}

\paragraph{Aggregation methods:} We categorise the methods into \textit{hard label}-based and \textit{soft label}-based.

\begin{enumerate}
    \item Hard labels: These methods result in a single label being assigned to each example, i.e., a one-hot vector.

\begin{itemize}[leftmargin=*]
    \item \textbf{Majority Vote (MV):} We take the most common first-choice label across annotators. Where there is no consensus, we chose a label at random from those annotated.
    \item \textbf{Majority Vote with Confidence (MVC):} Instead of treating each annotation equally, we weight the first-choice class by the annotator’s reported confidence, normalized into the range $[0,1]$. The class with the greatest confidence-weighted count among the annotators is chosen.
    \item \textbf{Majority Vote with Confidence and Second-choice (MVC2):} The same as MVC, but we also add in the second-choice label discounted by 2 times the normalised confidence score.
\end{itemize}

\item Soft labels: These methods result in a categorical distribution over labels.

\begin{itemize}[leftmargin=*]
    \item \textbf{Soft Vote (SV), SV with Confidence (SVC), and SVC and Second-choice (SVC2):} These three methods work the same as the corresponding hard label methods described above. However, instead of returning a single label, we return the distribution over labels computed by normalising the (weighted) counts into the range $[0,1]$.
    \item \textbf{Dawid-Skene (DS):} This method computes a distribution over labels by estimating a probabilistic graphical model of annotation errors in order to discover the underlying true label \cite{dawid1979maximum}.
    \item \textbf{Bayesian Soft Vote (BSV):} This method was developed specifically for cases such as ours where annotators provide a confidence score. While each annotator chooses their confidence from the same 1-5 Likert scale, their individual perceptions of what each step on the scale means may differ from one another. To account for this, Bayesian Soft Vote recalibrates each annotator’s confidence scores according to their level of agreement with other annotators by estimating a probabilistic graphical model \cite{wu-etal-2023-dont}.
\end{itemize}

\end{enumerate}

\noindent
For the methods that use the second-choice annotations, if the annotators choose \textit{Highly Uncertain} as their second-choice, we uniformly redistribute the probability mass remaining after the first-choice annotation to all other labels.

\paragraph{Comparison of Label Aggregation Methods}

\begin{figure}[h!]
    \centering
    \includegraphics[width=0.7\linewidth]{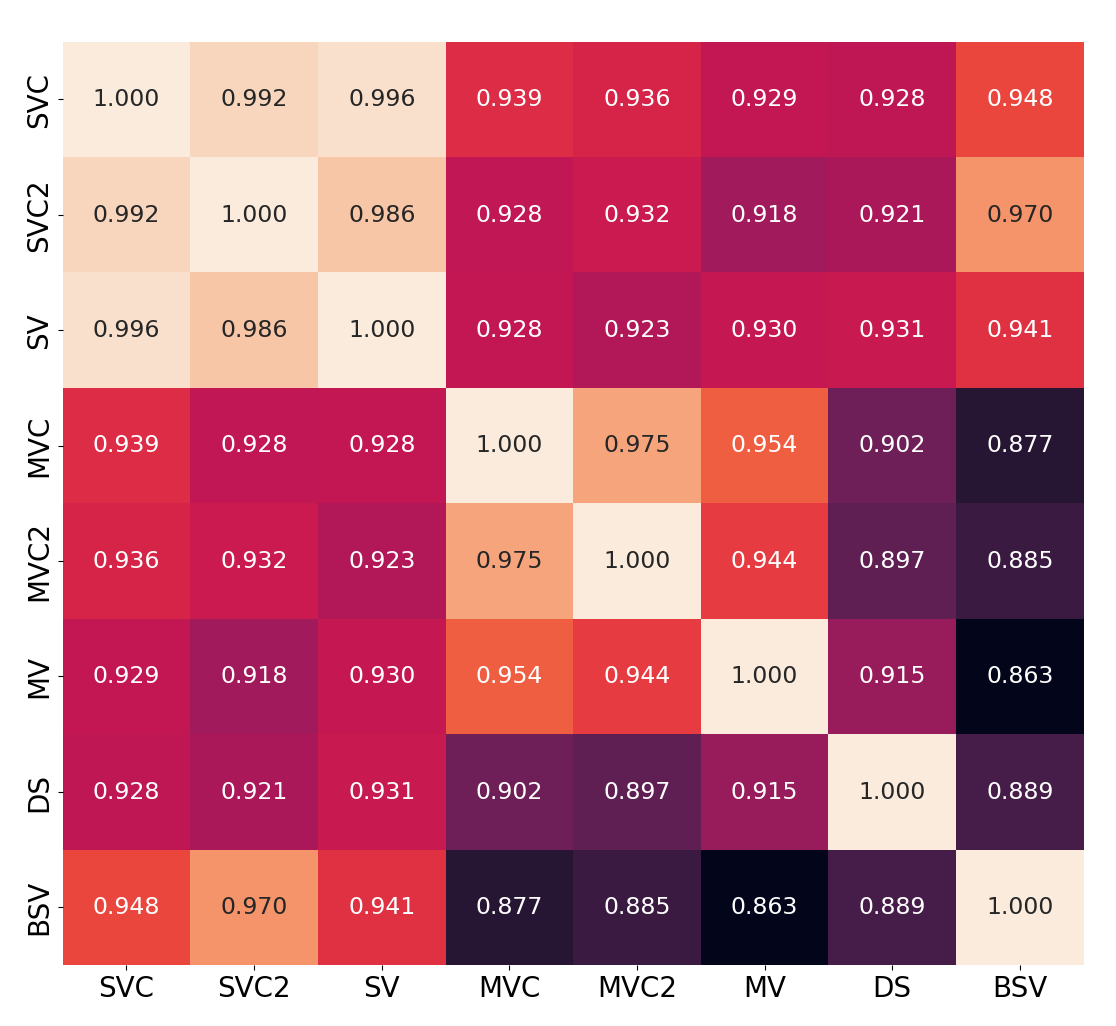}
    \caption{Cosine agreements among label aggregation methods, ordered from top-to-bottom according to the average overall agreement with the other methods.}
    \label{fig:aggregation_agreements}
\end{figure}

We compare the above label aggregation methods by computing the cosine agreement between the aggregated labels obtained using each pair of methods. As shown in \cref{fig:aggregation_agreements}, the overall the agreement between aggregation methods is high, with the lowest agreement being only 0.863. Note that a relatively low agreement (e.g., BSV) is not necessarily negative, as it simply shows that the method encodes information differently than the others, and there is not necessarily one ``best'' way to use the annotations. 


\section{Experimental Setups}

All the experiments were run on a single Nvidia A100 GPU with 40GB of VRAM. 

\paragraph{Synthetic Data Generation} We use the following prompt: \textit{Given a source tweet, generate 10 different replies in \{language\_name\} that \{stance\} the source tweet. Number each reply and put it on a new line. The source tweet is:\{source\_tweet\}.} We use sampling-based decoding, with temperature as 0.3.

\paragraph{Zero-shot ICL}
We use greedy search in decoding to ensure reproducibility. In the following prompt templates, \textbf{target\_input} denotes “\textit{Source tweet: \{source\_tweet\}. Reply tweet: \{reply\_tweet\}}”, and \textbf{task\_instruction} is “\textit{Source tweet: \{source\_tweet\}. Reply tweet: \{reply\_tweet\}. Determine the stance of the reply tweet towards the source tweet. The possible stance labels are “support", “deny", “query", or “comment". Answer with the stance label first, before any explanation. Definitions of the stance labels follow. “support": the reply tweet agrees with the source tweet. “deny" the reply disagrees with the source tweet. “query": the reply tweet asks for more information regarding the source tweet. “comment": the reply tweet does not take a clear stance towards the source tweet.}” 

\begin{itemize}
    \item \textbf{Baseline}: \textit{\{target\_input\}\{task\_instruction\}} with target inputs in original languages.

    \item \textbf{translate-input}: \textit{\{target\_input\}\{task\_instruction\}} with target inputs translated into English.

    \item \textbf{align example}: we use the following prompt: \textit{Use the following pairs of \{language name\} texts and their English translation to help you understand \{language name\}. Example \{num\_i\}: \{language name\}: \{tweet\}. English: \{English tweet\}. Now based on your understanding, answer the question below. \{target\_input\}\{task\_instruction\}}.
    
\end{itemize}

\paragraph{Few-Shot ICL} We use greedy search in decoding to ensure reproducibility.

\begin{itemize}
    \item \textbf{demo-en}: We use the following prompt: \textit{\{task\_instruction\} Example \{num\_i\}: Source tweet: \{source\_tweet\}. Reply tweet: \{reply\_tweet\}. Stance is: \{stance\}. Now complete the following example and answer with the stance label first before any explanation. \{target\_input\} Stance is}

    \item \textit{demo-translate}: We use the same prompt template as above, while the examples are translated into target non-English.

    \item \textit{demo-align}: \textit{\{task\_instruction\} Example \{num\_i\}: Source tweet: \{source\_tweet\}. \{language name\} translation is: {source tweet translation}. Reply tweet: \{reply\_tweet\}. \{language name\} translation is: \{reply tweet translation\}. Stance is: \{stance\}. Now complete the following example and answer with the stance label first before any explanation. \{target\_input\} Stance is}
    
\end{itemize}

\paragraph{Fine-Tuning MLMs} We use AdamW optimizer and search batch size from [16, 32] and learning rate from [1e-5,5e-5,1e-6,2e-6]. The optimal hyper-parameters are determined based on $wF2$ on RumourEval 2019 validation set. The number of training epochs is set as 5.

\section{Experimental Results}

\begin{table*}[h]
\centering
\scalebox{0.9}{
\begin{tabular}{lc|ccccccccc}
\toprule
&&CS&DE&ES&FR&HI&PL&PT&RU&EN\\
\midrule
\textit{XLM-R(Qwen)}&1st&0.3976&0.4315&0.4718&0.4929&0.3870&0.3933&0.4556&0.3133&0.4685\\
&2nd&0.4676&0.5198&0.5210&0.5700&0.4327&0.4840&0.5281&0.4321&0.5572\\
\midrule
\textit{XLM-R(Deepseek)}&1st&0.4769&0.4962&0.5181&0.5156&0.4490&0.4197&0.4907&0.3342&0.5426\\
&2nd&0.5551&0.5843&0.5724&0.6096&0.4905&0.5533&0.5645&0.4425&0.6404\\
\midrule
\textit{XLM-R(Llama)}&1st&0.4282&0.4590&0.4147&0.4050&0.3350&0.3764&0.3794&0.3500&0.4287\\
&2nd&0.4895&0.5304&0.4599&0.4790&0.3838&0.4552&0.4485&0.4378&0.5154\\
\midrule
\textit{XLM-R(Gemma)}&1st&0.3838&0.4072&0.4642&0.4765&0.4165&0.3829&0.4564&0.3105&0.4780\\
&2nd&0.4403&0.4985&0.5152&0.5460&0.4665&0.4865&0.5167&0.4149&0.5621\\
\midrule
\textit{XLM-R(Mistral)}&1st&0.4120&0.4510&0.4808&0.4903&0.4178&0.3658&0.4443&0.3476&0.4658\\
&2nd&0.4777&0.5397&0.5357&0.5787&0.4733&0.4879&0.5217&0.4778&0.5662\\
\midrule
\textit{XLM-T(Qwen)}&1st&0.4338&0.4528&0.5371&0.5010&0.4666&0.3899&0.5121&0.3797&0.4973\\
&2nd&0.4882&0.5244&0.5724&0.5630&0.5106&0.4803&0.5709&0.4884&0.5830\\
\midrule
\textit{XLM-T(Deepseek)}&1st&0.4343&0.4910&0.4629&0.4855&0.3996&0.3469&0.4357&0.2716&0.4957\\
&2nd&0.5208&0.5996&0.5223&0.5791&0.4574&0.4896&0.5183&0.3758&0.5861\\
\midrule
\textit{XLM-T(Llama)}&1st&0.4108&0.4559&0.4869&0.4807&0.3256&0.4255&0.4712&0.4223&0.4357\\
&2nd&0.4737&0.5251&0.5432&0.5566&0.3776&0.5460&0.5298&0.5559&0.5298\\
\midrule
\textit{XLM-T(Gemma)}&1st&0.4604&0.5174&0.5454&0.5350&0.4660&0.4737&0.5295&0.4441&0.5596\\
&2nd&0.5106&0.5931&0.5815&0.5931&0.4916&0.5648&0.5925&0.5497&0.6312\\
\midrule
\textit{XLM-T(Mistral)}&1st&0.3564&0.3653&0.4455&0.4329&0.4012&0.2944&0.4179&0.2264&0.4414\\
&2nd&0.4249&0.4770&0.4981&0.5224&0.4493&0.4015&0.4873&0.3257&0.5417\\
\bottomrule
\end{tabular}
    }
    \caption{Fine-tuning MLM with synthetic data, evaluated on the first choice and second choice labels.}
    \label{tab:ft-syn-full}
\end{table*}

\begin{table*}[h!]
\centering
\scalebox{0.7}{
\begin{tabular}{lc|ccccccccc}
\toprule
&&CS&DE&ES&FR&HI&PL&PT&RU&EN\\
\midrule
\textit{baseline(Qwen)} & 1st & 0.2773&0.2056&0.2672&0.3318&0.2247&0.2390&0.3278&0.2483&0.3741\\
&2nd&0.3478&0.3166&0.3053&0.4280&0.2653&0.3830&0.4103&0.3215&0.4803\\
\midrule
\textit{translate-input (Qwen)}&1st &0.3292&0.2475&0.3744&0.3600&0.3319&0.2664&0.3509&0.3316&-\\
& 2nd&0.4160&0.3643&0.4214&0.4650&0.4012&0.3487&0.4164&0.4032&-\\
\midrule
\textit{align example (Qwen)}&1st&0.2725&0.2413&0.3465&0.3697&0.2943&0.2491&0.3743&0.3082&-\\
&2nd&0.3417&0.3606&0.3910&0.4860&0.3517&0.3794&0.4433&0.3946&-\\
\midrule
\textit{baseline(Mistral)} & 1st &0.3847&0.3324&0.4376&0.4378&0.3316&0.3618&0.4663&0.4426&0.5032\\
&2nd&0.4695&0.4458&0.4970&0.5489&0.3906&0.4605&0.5466&0.5470&0.6208\\
\midrule
\textit{translate-input (Mistral)}&	1st &0.4289&0.3952&0.4843&0.4547&0.4360&0.3898&0.4631&0.4472&-\\
&2nd&0.5114&0.5065&0.5425&0.5706&0.5023&0.4882&0.5416&0.5532&-\\
\midrule
\textit{align example (Mistral)}&1st &0.4310&0.3644&0.4308&0.4603&0.3722&0.3894&0.4473&0.4445&-\\
&2nd&0.5079&0.4650&0.4879&0.5586&0.4276&0.5007&0.5367&0.5552&-\\
\midrule
\textit{baseline(Gemma)} & 1st &0.5391&0.5066&0.5893&0.5657&0.4830&0.5178&0.5535&0.4849&0.5702\\
&2nd&0.6120&0.6431&0.6491&0.6590&0.5351&0.6346&0.6190&0.5938&0.6758\\
\midrule
\textit{translate-input (Gemma)}&1st &0.6034&0.5298&0.6157&0.6016&0.5508&0.5099&0.5908&0.5368&-\\
&2nd&0.6540&0.6571&0.6699&0.6912&0.6027&0.6067&0.6580&0.6583&-\\
\midrule
\textit{align example (Gemma)}&1st &0.6412&0.6158&0.6610&0.6504&0.5754&0.5558&0.6092&0.5564&-\\
&2nd&0.6986&0.7124&0.7069&0.7313&0.6226&0.6428&0.6700&0.6682&-\\
\midrule
\textit{baseline(Deepseek)} & 1st &0.3391&0.3450&0.4161&0.4100&0.3097&0.2884&0.3910&0.3186&0.4446\\
&2nd&0.4155&0.4438&0.4624&0.4903&0.3626&0.3966&0.4435&0.4051&0.5486\\
\midrule
\textit{translate-input (Deepseek)}&1st &0.3674&0.3696&0.3886&0.3929&0.3470&0.3339&0.3663&0.3370&-\\
& 2nd&0.4481&0.4557&0.4384&0.4813&0.4019&0.4149&0.4207&0.4093&-\\
\midrule
\textit{align example (Deepseek)}&1st &0.3073&0.3259&0.3904&0.4196&0.2380&0.3184&0.3707&0.3087&-\\
&2nd&0.3720&0.4173&0.4374&0.4790&0.2793&0.4162&0.4234&0.4106&-\\
\midrule
\textit{baseline(Llama)} & 1st &0.2500&0.2434&0.3181&0.3008&0.2810&0.2176&0.3205&0.2971&0.3254\\
&2nd&0.2985&0.3023&0.3539&0.3499&0.3173&0.2878&0.3664&0.3695&0.3759\\
\midrule
\textit{translate-input (Llama)}&1st &0.3131&0.2651&0.3027&0.2912&0.2864&0.2811&0.3020&0.3290&-\\
&2nd&0.3595&0.3135&0.3371&0.3392&0.3183&0.3325&0.3423&0.3779&-\\
\midrule
\textit{align example (Llama)}&1st &0.1228&0.1533&0.1880&0.1986&0.1241&0.1384&0.1952&0.1394&-\\
&2nd&0.1425&0.1896&0.2160&0.2503&0.1448&0.1835&0.2358&0.1802&-\\
\bottomrule
\end{tabular}
    }
    \caption{Zero-shot ICL performance evaluated on the first choice and second choice labels.}
    \label{tab:zero-shot-full}
\end{table*}			

\begin{table*}[h]
\centering
\scalebox{0.9}{
\begin{tabular}{lc|cccccccccc}
\toprule
&&CS&DE&ES&FR&HI&PL&PT&RU&EN\\
\midrule
\textit{demo-en(Qwen)}&1st&0.3954&0.3833&0.4475&0.4657&0.3740&0.3520&0.4473&0.4881&0.5190&\\
&2nd&0.4697&0.4824&0.4997&0.5605&0.4270&0.4477&0.5108&0.5969&0.6169\\
\midrule
\textit{demo-translate(Qwen)}&1st&0.4147&0.3823&0.4604&0.4819&0.3995&0.3714&0.4558&0.4761&-&\\
&2nd&0.4868&0.4773&0.5172&0.5744&0.4628&0.4663&0.5125&0.5886&-\\
\midrule
\textit{demo-align(Qwen)}&1st&0.3850&0.3675&0.4223&0.4655&0.3610&0.3435&0.4205&0.4807&-&\\	
&2nd&0.4556&0.4597&0.4692&0.5415&0.4187&0.4325&0.4803&0.6028&-\\
\midrule
\textit{demo-en(Mistral)}&1st&0.4893&0.4556&0.5189&0.5054&0.4103&0.3969&0.5003&0.5405&0.5742&\\	
&2nd&0.5572&0.5635&0.5759&0.6118&0.4654&0.5014&0.5679&0.6508&0.6742\\
\midrule
\textit{demo-translate(Mistral)}&1st&0.4981&0.4530&0.5229&0.5151&0.4364&0.4025&0.5133&0.5274&-&\\
&2nd&0.5693&0.5635&0.5806&0.6220&0.4923&0.5074&0.5850&0.6386&-\\
\midrule
\textit{demo-align(Mistral)}&1st&0.4624&0.4271&0.4906&0.4779&0.3663&0.3752&0.4725&0.5167&-&\\
&2nd&0.5346&0.5225&0.5396&0.5747&0.4205&0.4726&0.5496&0.6073&-\\
\midrule
\textit{demo-en(Gemma)}&1st&0.6138&0.5948&0.6382&0.6260&0.5636&0.4969&0.6069&0.5542&0.6382&\\
&2nd&0.6931&0.6989&0.6897&0.7203&0.6130&0.6020&0.6711&0.6489&0.7418\\
\midrule
\textit{demo-translate(Gemma)}&1st&0.6109&0.5761&0.6416&0.6285&0.5472&0.5047&0.5998&0.5618&-&\\
&2nd&0.6917&0.6921&0.6896&0.7184&0.6007&0.6209&0.6657&0.6545&-\\
\midrule
\textit{demo-align(Gemma)}&1st&0.5587&0.5207&0.5873&0.5858&0.5057&0.4655&0.5646&0.5634&-&\\	
&2nd&0.6347&0.6331&0.6367&0.6705&0.5602&0.5686&0.6263&0.6483&-\\
\midrule
\textit{demo-en(Deepseek)}&1st&0.3830&0.4016&0.4589&0.4481&0.3662&0.3649&0.4227&0.4254&0.5036&\\
&2nd&0.4527&0.5049&0.5119&0.5453&0.4127&0.4644&0.4868&0.5031&0.5917\\
\midrule
\textit{demo-translate(Deepseek)}&1st&0.4392&0.4446&0.4846&0.4374&0.4077&0.3950&0.4359&0.4832&-&\\	
&2nd&0.5072&0.5308&0.5352&0.5245&0.4524&0.4874&0.4940&0.5514&-\\
\midrule
\textit{demo-align(Deepseek)}&1st&0.3641&0.3650&0.4411&0.3901&0.3637&0.3430&0.4100&0.4038&-&\\
&2nd&0.4351&0.4561&0.4909&0.4696&0.4075&0.4527&0.4733&0.4930&-\\
\midrule
\textit{demo-en(Llama)}&1st&0.4009&0.3579&0.3984&0.4149&0.3156&0.3111&0.3927&0.3896&0.4947&\\
&2nd&0.4679&0.4738&0.4447&0.5137&0.3647&0.4128&0.4666&0.4844&0.5869\\
\midrule
\textit{demo-translate(Llama)}&1st&0.4117&0.3609&0.4156&0.4257&0.2983&0.3025&0.4057&0.3884&-&\\	
&2nd&0.4886&0.4809&0.4614&0.5223&0.3484&0.3982&0.4808&0.4824&-\\
\midrule
\textit{demo-align(Llama)}&1st&0.2932&0.2349&0.3209&0.3019&0.2650&0.2173&0.3362&0.3528&-&\\
&2nd&0.3557&0.3246&0.3597&0.3977&0.3094&0.2848&0.4073&0.4413&-\\
\bottomrule
\end{tabular}
    }
    \caption{Few-shot ICL performance evaluated on the first choice and second choice labels.}
    \label{tab:few-shot-full}
\end{table*}

\begin{table*}[h]
\centering
\scalebox{0.9}{
\begin{tabular}{lc|ccccccccc|c}
\toprule
&&CS&DE&ES&FR&HI&PL&PT&RU&EN&RumourEval\\
\midrule
\textit{train-en(XLM-R)}&1st&0.3062&0.2435&0.2648&0.2783&0.2280&0.2307&0.2930&0.3231&0.3443& 0.4823\\
& 2nd & 0.3901&0.3525&0.2991&0.3675&0.2656&0.3284&0.3523&0.4253&0.4188&-\\
\midrule
\textit{train-translate(XLM-R)}&1st&0.2417&0.1895&0.2287&0.2250&0.2051&0.1453&0.2518&0.2354&0.2519&-\\
&2nd&0.2993&0.2606&0.2637&0.3114&0.2411&0.2742&0.3105&0.3005&0.3409&-\\
\midrule
\textit{train-en(XLM-T)}&1st&0.2557&0.2205&0.2916&0.2900&0.2095&0.2293&0.2934&0.3480&0.3445&0.4616\\
&2nd&0.3329&0.3112&0.3334&0.3928&0.2503&0.3345&0.3551&0.4626&0.4413&-\\
\midrule
\textit{train-translate(XLM-T)}&1st&0.1290&0.1564&0.1827&0.1958&0.1453&0.1635&0.2439&0.2271&0.2479&-\\
&2nd&0.1679&0.2266&0.2066&0.2557&0.1755&0.2212&0.2884&0.2770&0.3205&-\\
\bottomrule
\end{tabular}
    }
    \caption{Fine-tuning MLMs performance evaluated on the first choice and second choice labels. We also provide its performance on RumourEval 2019 test set for reference.}
    \label{tab:ft-full}
\end{table*}

\end{document}